\documentclass[10pt,twocolumn,letterpaper]{article}

\usepackage{cvpr}              % To produce the CAMERA-READY version
\usepackage{graphicx}
\usepackage{amsmath}
\usepackage{amssymb}
\usepackage{booktabs}

\usepackage[pagebackref,breaklinks,colorlinks]{hyperref}

\usepackage{float}
\usepackage{multirow}
\usepackage{amsmath}

\usepackage[capitalize]{cleveref}
\crefname{section}{Sec.}{Secs.}
\Crefname{section}{Section}{Sections}
\Crefname{table}{Table}{Tables}
\crefname{table}{Tab.}{Tabs.}

\usepackage{capt-of}
\usepackage{cuted}

\usepackage{appendix}

\def\cvprPaperID{3281} % *** Enter the CVPR Paper ID here
\def\confName{CVPR}
\def\confYear{2023}

\usepackage{amssymb}% http://ctan.org/pkg/amssymb
\usepackage{pifont}% http://ctan.org/pkg/pifont
\newcommand{\cmark}{\ding{51}}%

\makeatletter
\newcommand{\printfnsymbol}[1]{%
	\textsuperscript{\@fnsymbol{#1}}%
}

\begin{document}
	
	\title{NeRFLiX: High-Quality Neural View Synthesis \\by Learning a Degradation-Driven Inter-viewpoint MiXer
	\vspace{-0.2in} }
	% NeRFLiX: High-Quality Neural View Synthesis by Learning a Degradation-Aware Inter-viewpoint MiXer

	\author{
	Kun Zhou\textsuperscript{1,2}\printfnsymbol{1} \quad Wenbo Li\textsuperscript{3}\thanks{Equal contribution}  \quad Yi Wang\textsuperscript{4} \\
	Tao Hu\textsuperscript{3} \quad Nianjuan Jiang\textsuperscript{2} \quad Xiaoguang Han\textsuperscript{1}  \quad Jiangbo Lu\textsuperscript{2}\thanks{Corresponding author}  \\
	${^1}$SSE, CUHK-Shenzhen, \quad $^{2}$SmartMore Corporation 
	\quad $^{3}$CUHK \quad $^{4}$Shanghai AI Laboratory\\
	{\tt\small kunzhou@link.cuhk.edu.cn},  {\tt\small\{wenboli,taohu\}@cse.cuhk.edu.hk}\\
	{\tt\small\ hanxiaoguang@cuhk.edu.cn},{\tt\small \{jiangbo.lu,wygamle\}@gmail.com} \\
	\vspace{-0.8in}
}

%	\author{First Author\\
%		Institution1\\
%		Institution1 address\\
%		{\tt\small firstauthor@i1.org}
%		% For a paper whose authors are all at the same institution,
%		% omit the following lines up until the closing ``}''.
%		% Additional authors and addresses can be added with ``\and'',
%		% just like the second author.
%		% To save space, use either the email address or home page, not both
%		\and
%		Second Author\\
%		Institution2\\
%		First line of institution2 address\\
%		{\tt\small secondauthor@i2.org}
%	}

	\maketitle	

	\begin{strip}\centering
		\includegraphics[width=1.0\linewidth]{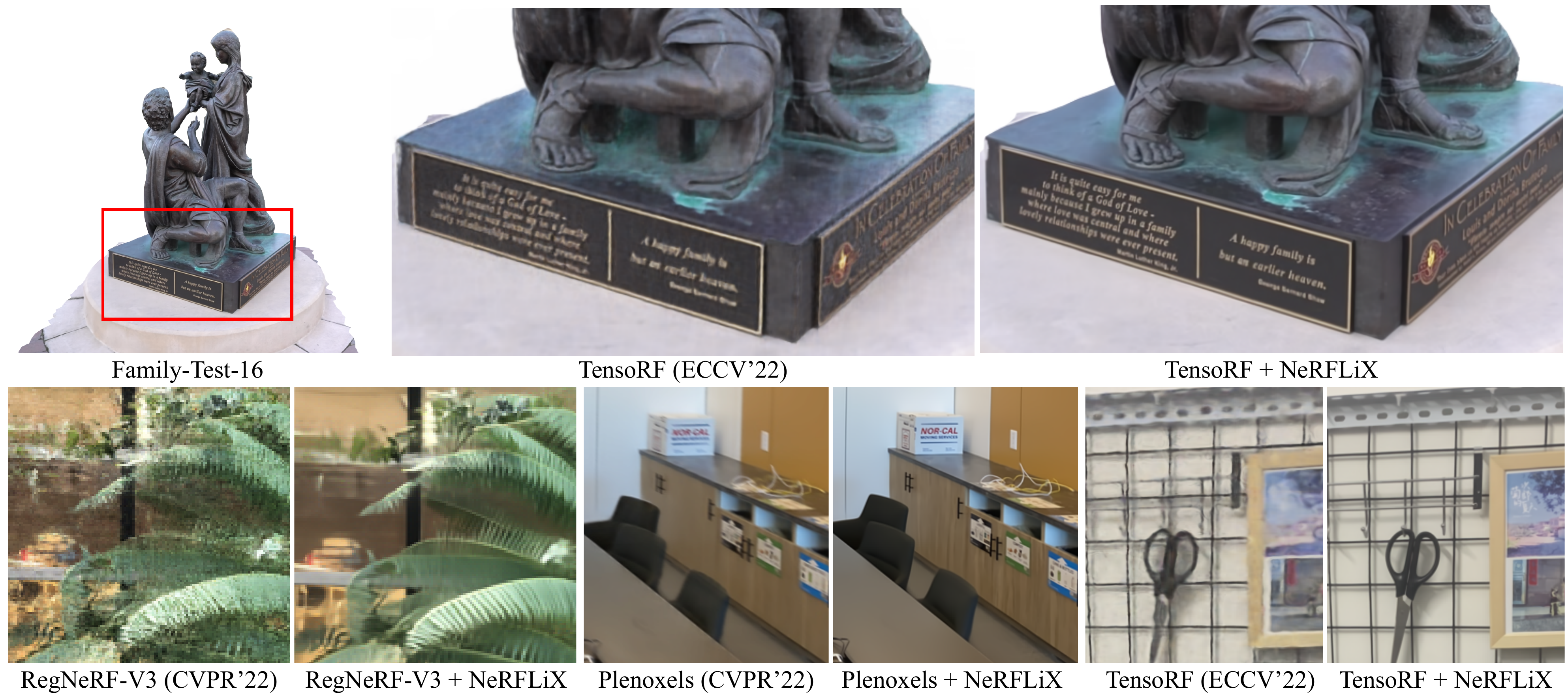}
		\captionof{figure}{We propose NeRFLiX, a general NeRF-agnostic restorer that is capable of improving neural view synthesis quality. The first example is from Tanks and Temples~\cite{knapitsch2017tanks}, the second/third examples are from LLFF~\cite{mildenhall2019local}, and the last one is a user scene captured by a mobile phone. RegNeRF-V3~\cite{Niemeyer2021Regnerf} means the model trained with three input views.}
		\label{fig:teasing}
	\end{strip}
	\vspace{-0.3in}
	%%%%%%%%% ABSTRACT
	\begin{abstract}
		Neural radiance fields~(NeRF) show great success in novel view synthesis. However, in real-world scenes, recovering high-quality details from the source images is still challenging for the existing NeRF-based approaches, due to the potential imperfect calibration information and scene representation inaccuracy. Even with high-quality training frames, the synthetic novel views produced by NeRF models still suffer from notable rendering artifacts, such as noise, blur, etc. Towards to improve the synthesis quality of NeRF-based approaches, we propose NeRFLiX, a general NeRF-agnostic restorer paradigm by learning a degradation-driven inter-viewpoint mixer. Specially, we design a NeRF-style degradation modeling approach and construct large-scale training data, enabling the possibility of effectively removing NeRF-native rendering artifacts for existing deep neural networks. Moreover, beyond the degradation removal, we propose an inter-viewpoint aggregation framework that is able to fuse highly related high-quality training images, pushing the performance of cutting-edge NeRF models to entirely new levels and producing highly photo-realistic synthetic views. Our project page is available at \url{https://redrock303.github.io/nerflix/}.
		
	\end{abstract}

	% \begin{figure}[t]
	% 	\centering
	% 	%\vspace{-0.15in}
	% 	\includegraphics[width=1.0\columnwidth]{figures/teasing1.png} 
	
	% 	\caption{We illustrate the effects of imperfect camera poses. To this end, we train two TensoRF~\cite{tensorf} models with perfect and noisy calibrated camera parameters (dubbed "TensoRF-Clean" and "TensoRF-Noise") on ``Lego" and ``Drums" from LLFF Synthetic~\cite{mildenhall2020nerf}. The noisy camera poses are obtained by applying random 3D transformation on the precise rotation matrix. (a-b) refer to the rendering frames by TensoRF-Noise and TensoRF-Clean, and (c) is the ground truth.
	% 	} 
	
	% 	\label{fig:teasing1}
	% \end{figure} %
	\vspace{-0.2in}
	%%%%%%%%% BODY TEXT
	\section{Introduction}
	\label{sec:intro}
	% describe 
	
	Neural radiance fields~(NeRF) can generate photo-realistic images from new viewpoints, playing a heated role in novel view synthesis. 
	In light of NeRF's~\cite{mildenhall2020nerf} success, numerous approaches~\cite{cole2021differentiable,tancik2022block,barron2021mip,yang2022recursive,pumarola2021d,martin2021nerf,wang2021ibrnet,mildenhall2022nerf,guo2022nerfren,mueller2022instant,chen2022aug,park2021nerfies,wang2021nerf,wu2022dof,wang2022fourier} along these lines have been proposed, continually raising the performance to greater levels. In fact,
	one prerequisite of NeRF is the precise camera settings of the taken photos for training ~\cite{lin2021barf,jeong2021self,wu2022dof}. However, accurately calibrating camera poses is exceedingly difficult in practice. Contrarily, the shape-radiance co-adaption issue~\cite{zhang2020nerf++} reveals that while the learned radiance fields can perfectly explain the training views with inaccurate geometry, they poorly generalize to unseen views. On the other hand, the capacity to represent sophisticated geometry, lighting, object materials, and other factors is constrained by the simplified scene representation of NeRF~\cite{zhang2021nerfactor,guo2022nerfren,zhang2022modeling}.
	On the basis of such restrictions, advanced NeRF models may nonetheless result in \textit{notable artifacts} (such as blur, noise, detail missing, and more), which we refer to as \textit{NeRF-style degradations} in this article and are shown in Fig.~\ref{fig:teasing}. 
	
	To address the aforementioned limitations, numerous works have been proposed. For example, some studies, including~\cite{yariv2020multiview,jeong2021self,wang2021nerf,zhang2022vmrf}, jointly optimize camera parameters and neural radiance fields to refine camera poses as precisely as possible in order to address the camera calibration issue. Another line of works~\cite{zhang2021nerfactor,guo2022nerfren,zhang2022modeling,zhang2021physg} presents physical-aware models that simultaneously take into account the object materials and environment lighting, as opposed to using MLPs or neural voxels to implicitly encode both the geometry and appearance. 
	To meet the demands for high-quality neural view synthesis, one has to carefully examine all of the elements when building complex inverse rendering systems. In addition to being challenging to optimize, they are also not scalable for rapid deployment with hard re-configurations in new environments. Regardless of the intricate physical-aware rendering models, \textit{is it possible to design a practical NeRF-agnostic restorer to directly enhance synthesized views from NeRFs}?
	
	In the low-level vision, it is critical to construct large-scale paired data to train a deep restorer for eliminating real-world artifacts~\cite{zhang2021designing,wang2021real}.  When it comes to NeRF-style degradations, there are two challenges: (1) sizable paired training data; (2) NeRF degradation analysis. First, it is unpractical to gather \textit{large-scale} training pairs (more specifically, raw outputs from well-trained NeRFs and corresponding ground truths). Second, the modeling of NeRF-style degradation has received little attention. Unlike real-world images that generally suffer from JPEG compression, sensor noise, and motion blur, the NeRF-style artifacts are complex and differ from the existing ones. As far as we know, \textbf{no} previous studies have ever investigated NeRF-style degradation removal which effectively leverages the ample research on image and video restoration.
	% Without in-depth exploiting those NeRF-specific artifacts and sizable paired training data, the NeRF-specific image restoration still under-exploited.
	% Those challenges explain why little effort has been put into applying image restoration models for NeRF-style artifact removal.
	
	% Without in-depth exploiting those NeRF-specific artifacts and sizable paired training data, existing image restoration frameworks fail to handle those challenging rendering degradations.
	In this work, we are motivated to have the \textit{first} study on the feasibility of simulating large-scale NeRF-style paired data, opening the possibility of training a NeRF-agnostic restorer for improving the NeRF rendering frames.
	To this end, we present a novel degradation simulator for typical NeRF-style artifacts (e.g., rendering noise and blur) considering the NeRF mechanism.  We review the overall NeRF rendering pipeline and discuss the typical NeRF-style degradation cases. Accordingly, we present three basic degradation types to simulate the real rendered artifacts of NeRF synthetic views and empirically evaluate the distribution similarity between real rendered photos and our simulated ones. The feasibility of developing NeRF-agnostic restoration models has been made possible by constructing a sizable dataset that covers a variety of NeRF-style degradations, over different scenes. 
	% we demonstrate how state-of-the-art deep neural works can be used to eliminate NeRF visual artifacts. 
	
	%  Benefiting from our simulated dataset, we find the state-of-the-art image restoration framework is useful to eliminate NeRF visual artifacts. 
	% However, relying entirely on a low-quality rendered frame still makes it difficult to restore photo-realistic results.
	Next, we show the necessity of our simulated dataset and demonstrate that existing state-of-the-art image restoration frameworks can be used to eliminate NeRF visual artifacts. Furthermore,
	we notice, in a typical NeRF setup, neighboring high-quality views come for free, and they serve as potential reference bases for video-based restoration with a multi-frame aggregation and fusion module.
	However, this is not straightforward because NeRF input views are taken from a variety of very different angles and locations, making the estimation of correspondence quite challenging. To tackle this problem, we propose a degradation-driven inter-viewpoint ``mixer" that progressively aligns image contents at the pixel and patch levels. In order to maximize efficiency and improve performance, we also propose a fast view selection technique to only choose the most pertinent reference training views for aggregation, as opposed to using the entire NeRF input views.

	% Last but not least, we demonstrate our approach can be employed to accelerate the training of various SOTA NeRFs, while still delivering on-par or higher performance. For instance, we show that Plenoxels~\cite{fridovich2022plenoxels} models can be early terminated at 10 minutes~(2x time faster than the original Plenoxels) while keeping their performance with the help of our restorer.
	
	In a nutshell, we present a NeRF-agnostic restorer (termed NeRFLiX) which learns a degradation-driven inter-viewpoint mixer. As illustrated in Fig.~\ref{fig:teasing}, given NeRF synthetic frames with various rendering degradations, NeRFLiX successfully restores high-quality results. Our contributions are summarized as
	
	\begin{itemize}
		
		\item 
		\noindent\textbf{{Universal enhancer for NeRF models.}} NeRFLiX is powerful and adaptable, removing NeRF artifacts and restoring clearly details,  pushing the performance of cutting-edge NeRF models to entirely new levels.
		
		\item 
		\noindent\textbf{NeRF rendering degradation simulator.} We develop a NeRF-style degradation simulator~(NDS), constructing massive amounts of paired data and aiding the training of deep neural networks to improve the quality of NeRF-rendered images.
		\item 
		\noindent\textbf{Inter-viewpoint mixer.} Based on our constructed NDS, we further propose an inter-viewpoint baseline that is able to \textit{mix} high-quality neighboring views for more effective restorations.
		
		\item \noindent\textbf{{Training time acceleration.}} We show how NeRFLiX makes it possible for NeRF models to produce even \textit{better} results with a \textit{50$\%$} reduction in training time.
	\end{itemize}
	% \end{enumerate}
	
	\section{Related Works}
	\label{sec:relatedwork}
	\vspace{-0.05in}
	\noindent\textbf{NeRF-based novel view synthesis.}
	NeRF-based novel view synthesis has received a lot of attention recently and has been thoroughly investigated. For the first time, Mildenhall~\etal ~\cite{mildenhall2020nerf} propose the neural radiance field to implicitly represent static 3D scenes and synthesize novel views from multiple posed images. Inspired by their successes, a lot of NeRF-based models~\cite{cole2021differentiable,tancik2022block,barron2021mip,yang2022recursive,pumarola2021d,martin2021nerf,wang2021ibrnet,hu2022efficientnerf,mildenhall2022nerf,xiang2021neutex,guo2022nerfren,mueller2022instant,suhail2022light,kurz2022adanerf,johari2022geonerf,ichnowski2021dex,lin2021efficient,chen2021mvsnerf,deng2022compressing,zhang2021physg,wang2022nerf,rebain2021derf,zhang2022fast} have been proposed. For example, point-NeRF~\cite{xu2022point} and DS-NeRF~\cite{deng2022depth} incorporate sparse 3D point cloud and depth information for eliminating the geometry ambiguity for NeRFs, achieving more accurate/efficient 3D point sampling and better rendering quality. Plenoxels~\cite{fridovich2022plenoxels}, TensoRF~\cite{tensorf}, DirectVoxGo~\cite{sun2022direct}, FastNeRF~\cite{garbin2021fastnerf}, Plenoctrees~\cite{yu2021plenoctrees}, KiloNeRF~\cite{reiser2021kilonerf},  and Mobilenerf~\cite{chen2022mobilenerf},  aim to use various advanced technologies to speed up the training or inference phases. Though these methods have achieved great progress, due to the potential issue of inaccurate camera poses, simplified pinhole camera models as well as scene representation inaccuracy, they still suffer from rendering artifacts of the predicted novel views. 
	
	\vspace{0.03in}
	\noindent\textbf{Degradation simulation.}
	Since no existing works have explored the NeRF-style degradation cases, we will overview the real-world image restoration works that are most related to ours. The previous image/video super-resolution approaches~\cite{li2022best,dong2015image,zhang2018residual,zhou2021revisiting,wang2019edvr,li2020lapar,wang2018esrgan,liang2021swinir,zhang2018residual,yu2021path} typically follow a fix image degradation type~(e.g., blur, bicubic/bilinear down-sampling). Due to the large domain shift between the real-world and simulated degradations, the earlier image restoration methods~\cite{li2020lapar,li2019feedback,zhang2018residual,zhang2020deep} generally fail to remove complex artifacts of the real-world images. In contrast, BSRGAN~\cite{zhang2021designing} design a practical degradation approach for real-world image super-resolution. In their degradation process, multiple degradations are considered and applied in random orders, largely covering the diversity of real-world degradations. Compared with the previous works, BSRGAN achieves much better results quantitatively and qualitatively. Real-ESRGAN~\cite{wang2021real} develops a second-order degradation process for real-world image super-resolution. In this work, we propose a NeRF-style degradation simulator and construct a large-scale training dataset for modeling the NeRF rendering artifacts.
	
	\vspace{0.03in}
	\noindent\textbf{Correspondence estimation.}
	In the existing literature, the video restoration methods~\cite{yu2020joint,wang2018learning,teed2020raft,chan2021basicvsr++,cao2021vsrt} aim to restore a high-quality frame from multiple low-quality frames. To achieve this goal, cross-frame correspondence estimation is essential to effectively aggregate the informative temporal contents. Some works~\cite{yu2020joint,xue2019video,chan2021basicvsr++,chan2021basicvsr} explore building pixel-level correspondences through optical-flow estimation and perform frame-warping for multi-frame compensation. Another line of works~\cite{wang2019edvr,tian2020tdan,zhou2021revisiting} tries to use deformable convolution networks~(DCNs~\cite{dai2017deformable}) for adaptive correspondence estimation and aggregation. More recently, transformer-based video restoration models~\cite{liang2022vrt,cao2022vdtr} implement spatial-temporal aggregation through an attention mechanism and achieve promising performance. However, it is still challenging to perform accurate correspondence estimation between frames captured with very distinctive viewpoints.

	\section{Preliminaries}
	\begin{figure}[ht]
		\centering
		%\vspace{-0.15in}
		\includegraphics[width=0.95\columnwidth]{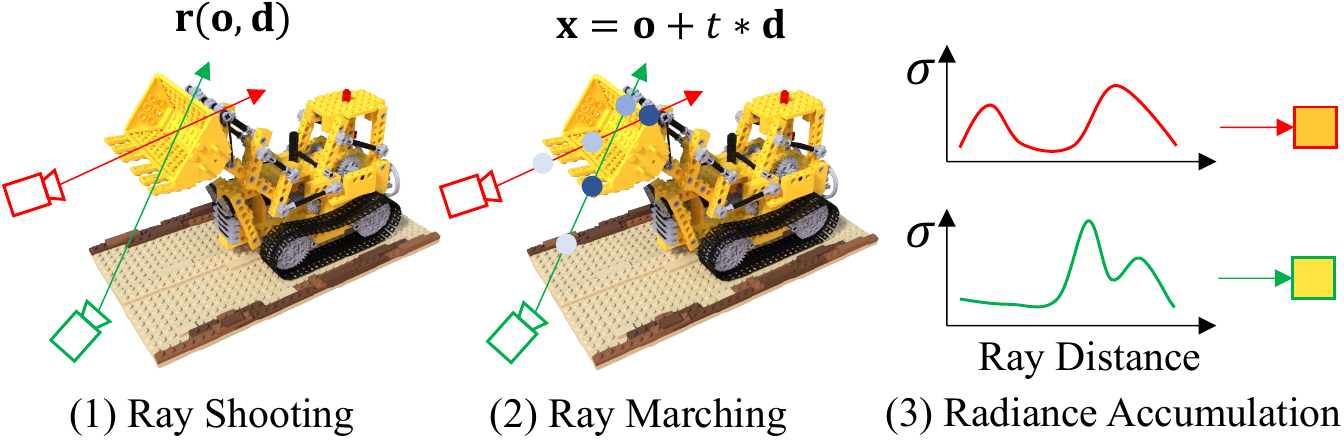} 
		\caption{A general illustration of NeRF-based novel view synthesis pipeline. Three main steps are involved: (1) ray shooting, (2) ray marching, and (3) radiance accumulation.
		} 
		\vspace{-0.1in}
		\label{fig:NeRFIll}
	\end{figure} % 
	In this section, we review the general pipeline of NeRF-based novel view synthesis and discuss potential rendering artifacts. As shown in Fig.~\ref{fig:NeRFIll}, three main steps are involved in the rendering: 
	% (1) ray shooting, (2) ray marching, and (3) radiance accumulation.
	(1) Ray Shooting.
	To render the color of a target pixel in a particular view, NeRF utilizes the camera's calibrated parameters~$\pi$ to generate a ray $\mathbf{r(o,d)}$ through this pixel, where $\mathbf{o},\mathbf{d}$ are the camera center and the ray direction.
	(2) Ray Marching. 
	A set of 3D points are sampled along the chosen ray as it moves across the 3D scene represented by neural radiance fields. The NeRF models encode a 3D scene and predict the colors and densities of these points.
	(3) Radiance Accumulation.
	The pixel color is extracted by integrating the predicted radiance features of the sampled 3D points.
	
	% In the training stage, NeRF creates neural radiance fields to represent 3D scenes under the supervision of the captured multi-view images.
	
	% \begin{figure*}[ht]
	% 	\centering
	% 	%\vspace{-0.15in}
	% 	\includegraphics[width=18cm]{figures/Method_overview.png} 
	
	% 	\caption{Overview of our NeRFLiX. (a) We gather a large-scale high-quality sequences. (b) The proposed NeRF degradation Simulation generates degraded images that are visually similar to real rendered images, (c) Our inter-viewpoint mixer takes the degraded frame and some overlapped clean frames for aggregation and outputs a high-quality result. 
	% 	} 
	
	% 	\label{fig:pipeline}
	% \end{figure*} % 
	
	\vspace{0.05in}
	\noindent\textbf{Discussion.} We can see that establishing a relationship between 2D photos and the 3D scene requires camera calibration. Unfortunately, it is very challenging to precisely calibrate the camera poses, leading to noisy 3D sampling. Meanwhile, some previous works~\cite{yariv2020multiview,jeong2021self,wang2021nerf,zhang2022vmrf} also raise other concerns, including the non-linear pinhole camera model~\cite{jeong2021self} and shape-radiance ambiguity~\cite{zhang2020nerf++}. Because of these inherent limitations, as discussed in Section~\ref{sec:intro}, NeRF models still synthesize unsatisfied novel test views.

	\begin{figure*}[ht]
		\centering
		%\vspace{-0.15in}
		\includegraphics[width=18cm]{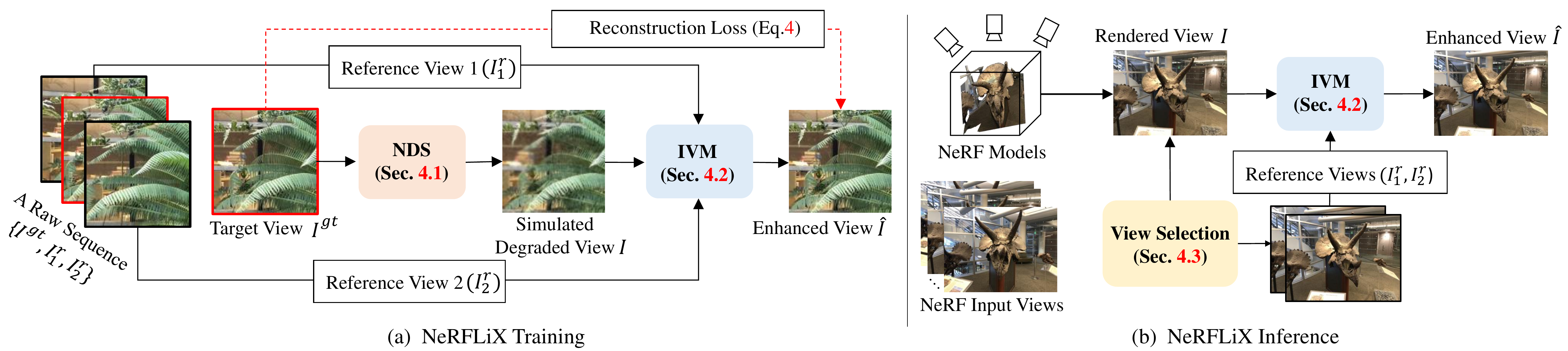} 
		% \includegraphics[width=1.0\columnwidth]{figures/t1.png} 
		% \caption{Overview of our NeRFLiX. (a) We gather a large-scale high-quality sequences. (b) The proposed NeRF degradation Simulation generates degraded images that are visually similar to real rendered images, (c) Our inter-viewpoint mixer takes the degraded frame and some overlapped clean frames for aggregation and outputs a high-quality result. 
		% } 
		\caption{Illustration of our proposed NeRFLiX. It consists of two essential modules: (1) NeRF degradation simulator that constructs paired training data $\{I,I_1^r,I_2^r|I^{gt}\}$ from a raw sequence $\{I^{gt},I_1^r,I_2^r\}$, (2) inter-viewpoint mixer trained on this simulated data is capable of restoring high-quality frames from NeRF rendered views. }
		\vspace{-0.18in}
		\label{fig:pipeline}
	\end{figure*} % 
	
	\section{Methodology}
	%	In this work, we propose a general NeRF enhancement framework that composes of NeRF-specific artifacts simulator for collecting clean/noisy paired data, 
	% In this work, we propose NeRFLiX, which learns a degradation-driven inter-viewpoints mixer. It composes of two key parts: (1) a NeRF-style artifacts simulator and (2) an inter-viewpoint mixer. 
	
	\noindent\textbf{Overview.}
	In this work, we present NeRFLiX, a general NeRF-agnostic restorer which employs a degradation-driven inter-viewpoint mixer to enhance novel view images rendered by NeRF models.
	It is made up of two essential components: a NeRF-style degradation simulator~(NDS) and an inter-viewpoint mixer~(IVM). As seen in Fig.~\ref{fig:pipeline}(a), during the training phase, we employ the proposed NDS to create large-scale paired training data, which are subsequently used to train an IVM for improving a NeRF-rendered view using two corresponding reference pictures (reference views). In the inference stage, as illustrated in Fig.~\ref{fig:pipeline}(b), IVM is adopted to enhance a rendered view by fusing useful information from the selected most relevant reference views.

	\vspace{-0.03in}
	\subsection{NeRF-Style Degradation Simulator~(NDS)}
	\label{sec:degraded}
	
	%  Since very limited \textit{paired} NeRF data is available and one has to collect large-scale well-posed scenes under different environments and train various NeRFs of each scene from scratch--it is infeasible to gather large-scale paired data. Here, we aim to design a practical NeRF degradation simulator to generate a large-scale training dataset that is not only visually but also statistically similar to the NeRF-rendered images.
	
	Due to the difficulties in gathering well-posed scenes under various environments and training NeRF models for each scene, it is infeasible to directly collect large amounts of \textit{paired} NeRF data for artifact removal.
	To address this challenge, motivated by BSRGAN~\cite{zhang2021designing},
	we design a general NeRF degradation simulator to produce a sizable training dataset that is visually and statistically comparable to NeRF-rendered images~(views).
	
	To begin with, we collect raw data from LLFF-T\footnote{the training parts of LLFF~\cite{mildenhall2019local}.} and Vimeo90K\cite{xue2019video} where the adjacent frames are treated as raw sequences. Each raw sequence consists of three images $\{I^{gt},I_1^r,I_2^r\}$: a target view $I^{gt}$ and its two reference views $\{I_1^r,I_2^r\}$. To construct the paired data from a raw sequence, we use the proposed NDS to degrade $I^{gt}$ and obtain a simulated degraded view $I$, as shown in Fig.~\ref{fig:pipeline}(a). 
	%Next, we go into further depth about our degrading system below.
	
\begin{figure}[t]
		\centering
		%\vspace{-0.15in}
		\includegraphics[width=0.95\columnwidth]{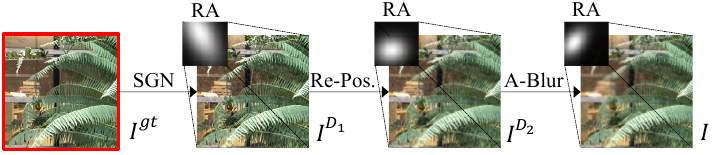} 
		
		\caption{Overview of our NDS pipeline: using our proposed degradations, we process a target view $I^{gt}$ to produce its simulated degraded view $I$. ``SGN", ``Re-Pos." and ``A-Blur" refer to the splatted Gaussian, re-positioning, anisotropic blur degradations, and ``RA" is the region adaptive strategy.
			\vspace{-0.15in}
		} 
		
		\label{fig:nds}
	\end{figure}

	%	In the following paragraphs, we will explain the main two degradation types in our NeRF-style degradation simulator: (1) diffused Gaussian noise, (2) an-isotropic blur.
	% Given a clean image $I_r^{gt}$, we aim to utilize the proposed NSDS to produce the degraded image $I_r$ with three main degradation types: (1) diffused Gaussian noise, (2)re-position and (3) an-isotropic blur.
	The degradation pipeline is illustrated in Fig~\ref{fig:nds}. We design three types of degradation for a target view $I^{gt}$: splatted Gaussian noise~(SGN), re-positioning~(Re-Pos.), and anisotropic blur~(A-Blur). It should be noted that there \textit{may be other models for such a simulation}, and we only utilize this route to evaluate and justify the feasibility of our idea.
	
	\vspace{0.05in}
	\noindent\textbf{Splatted Gaussian noise.}
	Although additive Gaussian noise is frequently employed in image/video denoising, NeRF rendering noise clearly differs. Rays that hit a 3D point will be re-projected within a nearby 2D area because of noisy camera parameters. As a result, the NeRF-style noise is dispersed over a 2D space. This observation led us to present a splatted Gaussian noise, which is defined as
	\begin{equation}
		I^{D1} = (I^{gt}  + n) \circledast g ,
		\vspace{-0.05in}
	\end{equation}
	where $n$ is a 2D Gaussian noise map with the same resolution as $I^{gt}$ and $g$ is an isotropic Gaussian blur kernel.
	
	\vspace{0.05in}
	\noindent\textbf{Re-positioning.}
	We design a re-positioning degradation to simulate ray jittering. We add a random 2D offset $\delta_i,\delta_j \in [-2,2]$ with probability 0.1 for a pixel at location $(i,j)$
	\vspace{-0.05in}
	\begin{equation}
		I^{D2}(i,j) = 
		\begin{cases}
			I^{D1}(i,j) &\text{if~~} p>0.1 \\
			I^{D1}(i+ \delta_i,j+\delta_j) &\text{else~~} p\leq0.1
		\end{cases}
		\vspace{-0.05in}
	\end{equation}
	where $p$ is uniformly distributed in $[0, 1]$.
	% Also, we smooth the image using two distinct isotropic Gaussian blur kernels to prevent excessive deterioration and combine the results.

	\vspace{0.03in}
	\noindent\textbf{Anisotropic blur.}
	Additionally, from our observation, NeRF synthetic frames also contain blurry contents. To simulate blur patterns, we use anisotropic Gaussian kernels to blur the target frame.
	% Incorrect calibration information also introduces multi-view inconsistency. Here is a situation, if rays sampled around object structures~(both geometric or textured) deviate from their ground truth, the re-projection operations will make blurry contents as verified in Figure~\ref{fig:NeRFNoise}. Following the previous real-world image super-resolution works, we employ anisotropic Gaussian blur kernels applied on the target frame to generate blur contents.

	\begin{figure}[h]
		\centering
		%\vspace{-0.15in}
		\includegraphics[width=1.0\columnwidth]{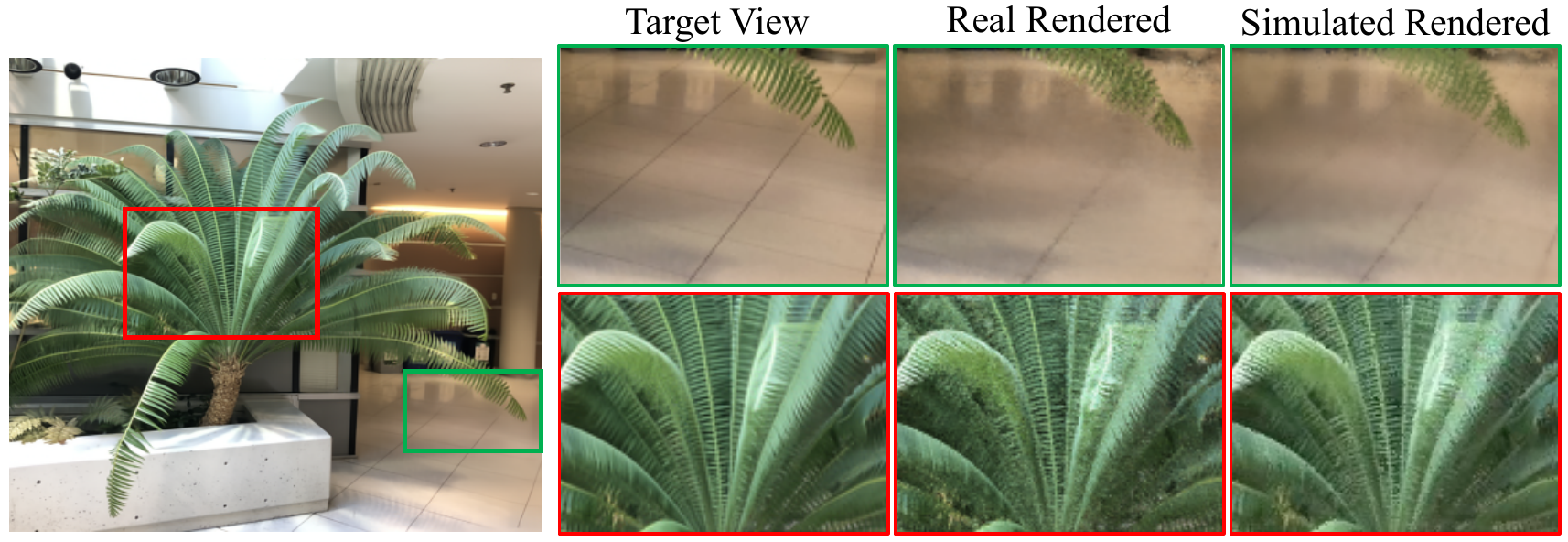} 
		
		\caption{A visual example of real and simulated rendered views.
			\vspace{-0.15in}
		} 
		\label{fig:synviz}
	\end{figure} % 
	%\paragraph{Region Adaptive Degradation}
	% Those areas that receive denser rays show higher rendering quality than those that are only shot by fewer rays. Therefore, different image areas illustrate varying degradation levels. To visually model this effect, we present a region-adaptive degradation strategy. To this end, each of the used degradation types will be performed repeatedly. For each time, a randomly 2D mask is applied to achieve spatial-varying effects. 
	\vspace{0.05in}
	\noindent\textbf{Region adaptive strategy.}
	Neural radiance fields are often supervised with unbalanced training views. As a result, given a novel view, the projected 2D areas have varying degradation levels. Thus, we carry out each of the employed degradations in a spatially variant manner. More specifically, we define a mask $M$ as a two-dimensional oriented anisotropic Gaussian~\cite{geusebroek2003fast}
	\begin{equation} 
		\vspace{-0.02in}
		M(i,j)= G(i-c_i, j-c_j; \sigma_i, \sigma_j, A),
		\vspace{-0.02in}
	\end{equation} 
	where $(c_i,c_j),(\sigma_i, \sigma_j)$ are the means and standard deviations and $A$ is an orientation angle. After that, we use the mask $M$ to linearly blend the input and output of each degradation, finally achieving region-adaptive degradations. As shown in Fig.~\ref{fig:synviz}, our simulated rendered views visually match the real NeRF-rendered ones. All the detailed settings of NDS are elaborated in our supplementary materials.
	
	At last, with our NDS, we can obtain a great number of training pairs, and each paired data consists of two high-quality reference views $\{I_1^{r},I_2^{r}\}$, a simulated degraded view $I$, and the corresponding target view $I^{gt}$. Next, we show how the constructed paired data $\{I,I_1^{r},I_2^{r}|I^{gt}\}$ can be used to train our IVM.

	% \paragraph{Discussion}
	% 	As illustrated in Fig.~\ref{fig:synviz}, our simulated image visually matches the real NeRF rendering image. Also, we conduct an extensive investigation of degradation strategies in the latter experiment section.
	
	\begin{figure}[t]
		\centering
		%\vspace{-0.15in}
		\includegraphics[width=1.0\columnwidth]{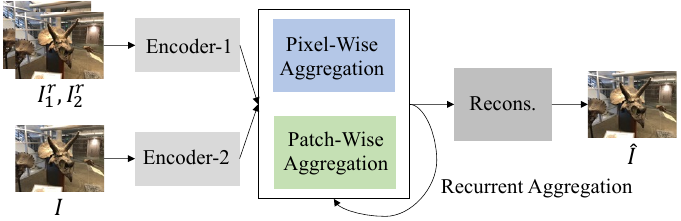} 
		
		\caption{The framework of our inter-viewpoint mixer.
		} 
		\vspace{-0.2in}
		\label{fig:ivm}
	\end{figure} % 
	
	\subsection{Inter-viewpoint Mixer~(IVM)}
	\label{sec:ivm}
	
	% \vspace{0.05in}
	\noindent\textbf{Problem formulation.} Given a degraded view $I$ produced by our NDS or NeRF models, we aim to extract useful information from its two high-quality reference views $\{I_1^r,I_2^r\}$ and restore an enhanced version $\hat I$.

	\vspace{0.03in}
	\noindent\textbf{IVM architecture.}
	For multi-frame processing, existing techniques either use optical flow~\cite{chan2021basicvsr,yu2020joint,wang2018learning} or deformable convolutions~\cite{dai2017deformable,wang2019edvr,liang2022vrt} to realize the correspondence estimation and aggregation for \textit{consistent} displacements. In contrast, NeRF rendered and input views come from very different angles and locations, making it challenging to perform precise inter-viewpoint aggregation.
	% For multi-frame processing, existing techniques either use optical flow~\cite{chan2021basicvsr,yu2020joint,wang2018learning} or deformable convolutions~\cite{dai2017deformable,wang2019edvr,liang2022vrt} to realize the correspondence estimation and aggregation for \textit{consistent} displacements. However, under a typical NeRF setup, it is challenging to carry out precise inter-viewpoint aggregation since rendered and input views have large viewpoint changes.
	% To achieve inter-viewpoint aggregation, we propose IVM, a hybrid recurrent inter-viewpoint ``mixer", to address this problem.
	
	To address this problem, we propose IVM, a hybrid recurrent inter-viewpoint ``mixer" that progressively fuses pixel-wise and patch-wise contents from two high-quality reference views, achieving more effective inter-viewpoint aggregation. There are three modules i.e., feature extraction, hybrid inter-viewpoint aggregation and reconstruction, as shown in Fig.~\ref{fig:ivm}. Two convolutional encoders are used in the feature extraction stage to process the degraded view $I$ and two high-quality reference views $\{I_1^r,I_2^r\}$, respectively. We then use inter-viewpoint window-based attention modules and deformable convolutions to achieve recurrent patch-wise and pixel-wise aggregation. Finally, the enhanced view ${\hat I}$ is generated using the reconstruction module under the supervision
	\vspace{-0.05in}
	\begin{equation}
		\begin{split}
			Loss = |{\hat I} - I^{gt}|, \text{where  } {\hat I} = f(I,I_1^r,I_2^r;\theta),
		\end{split}
		\vspace{-0.05in}
	\end{equation}
	where $\theta$ is the learnable parameters of IVM. The framework architecture is given in our supplementary materials.

	\subsection{View Selection}
	\begin{figure}[t]
		\centering
		%\vspace{-0.15in}
		\includegraphics[width=0.95\columnwidth]{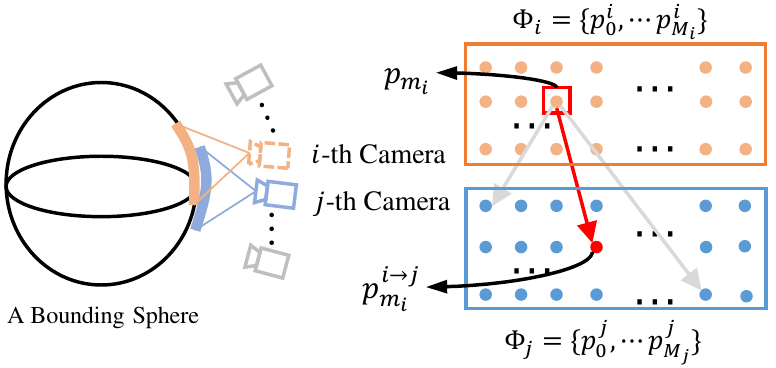} 
		
		\caption{Illustration of our view selection strategy. 
		} 
		\vspace{-0.2in}
		\label{fig:vm}
	\end{figure} % 
	\begin{table*}[t]
		\label{tab:nsds}
		\small
		\setlength{\tabcolsep}{3pt}
		\begin{subtable}[b]{0.46\textwidth}
			
			\begin{tabular}{l|c|c|c} %p{2cm}p{2cm}
				\hline
				Method                      & PSNR~(dB)${\color{red}\uparrow}$ &SSIM${\color{red}\uparrow}$ &LPIPS${\color{red}\downarrow}$   \\ \hline
				
				TensoRF~\cite{tensorf}  {\footnotesize (ECCV'22)}          			 &  26.73      		     & 0.839  &0.204  \\ 
				TensoRF~\cite{tensorf} + NeRFLiX                 &  \textbf{27.39} ($\color{red}\uparrow$ 0.66)         &   \textbf{0.867} &\textbf{0.149} \\  \hline \hline

				Plenoxels~\cite{fridovich2022plenoxels}  {\footnotesize(CVPR'22)}            			 &  26.29      		     & 0.839  &0.210  \\ 
				Plenoxels~\cite{fridovich2022plenoxels} + NeRFLiX                 &  \textbf{26.90} ($\color{red}\uparrow$ 0.61)          &   \textbf{0.864} &\textbf{0.156} \\  \hline \hline

				NeRF-mm~\cite{wang2021nerf}  {\footnotesize(ARXIV'21) }            			 & 22.98    		     & 0.655  &0.440  \\ 
				NeRF-mm~\cite{wang2021nerf} + NeRFLiX                 &  \textbf{23.38} ($\color{red}\uparrow$ 0.40)         &   \textbf{0.694} &\textbf{0.360} \\  \hline \hline
				
				NeRF~\cite{mildenhall2020nerf} {\footnotesize (ECCV'20)}           			 & 26.50    		     & 0.811  &0.250  \\ 
				NeRF~\cite{mildenhall2020nerf} + NeRFLiX                 &  \textbf{27.26}($\color{red}\uparrow$ 0.76)         &   \textbf{0.863} &\textbf{0.159} \\  \hline
			\end{tabular}
			\caption{Quantitative results on the LLFF under LLFF-P1.}
			\label{subtab:sotallff1}
		\end{subtable}
		\hfill
		\begin{subtable}[b]{0.46\textwidth}
			
			\begin{tabular}{l|c|c|c } %p{2cm}p{2cm}
				\hline
				Method                      & PSNR~(dB)${\color{red}\uparrow}$ &SSIM${\color{red}\uparrow}$ &LPIPS${\color{red}\downarrow}$   \\ \hline
				
				NLF~\cite{attal2022learning} {\footnotesize(CVPR'22)}           			 & 27.46     		     & 0.868  &0.136  \\ 
				NLF~\cite{attal2022learning} + NeRFLiX                 &  \textbf{28.19} ($\color{red}\uparrow$ 0.73)         &   \textbf{0.899} &\textbf{0.093} \\  \hline \hline
				RegNeRF-V3~\cite{Niemeyer2021Regnerf} {\footnotesize(CVPR'22)}    & 19.10		     & 0.587  &0.373  \\ 
				RegNeRF-V3~\cite{Niemeyer2021Regnerf} + NeRFLiX    &\textbf{19.68} ($\color{red}\uparrow$ 0.58)		     &\textbf{ 0.661}  &\textbf{0.260}  \\ \hline
				
				RegNeRF-V6~\cite{Niemeyer2021Regnerf} {\footnotesize(CVPR'22)}    &23.06		     & 0.759  &0.242  \\ 
				RegNeRF-V6~\cite{Niemeyer2021Regnerf} + NeRFLiX    &\textbf{23.90}($\color{red}\uparrow$ 0.84)		     & \textbf{0.815}  &\textbf{0.144}  \\ \hline
				
				RegNeRF-V9~\cite{Niemeyer2021Regnerf} {\footnotesize(CVPR'22)}    & 24.81     		     & 0.818  &0.196  \\ 
				RegNeRF-V9~\cite{Niemeyer2021Regnerf} + NeRFLiX    &\textbf{25.68} ($\color{red}\uparrow$ 0.87)		     &\textbf{ 0.863}  &\textbf{0.114 } \\ \hline
			\end{tabular}
			
			\caption{Quantitative results on the LLFF~\cite{mildenhall2019local} under LLFF-P2. RegNeRF-V3(6,9) takes 3(6,9) input views for training.}
			\label{subtab:sotallff2}
		\end{subtable}

		\caption{Quantitative analysis of our NeRFLiX on LLFF~\cite{mildenhall2019local}.}\label{tab:sotallff}
		\vspace{-0.15in}
		\label{tab:sota1}
	\end{table*}
	In the inference stage, for a NeRF-rendered view $I$, our IVM produces an enhanced version by aggregating contents from two neighboring high-quality views. But, multiple input views are available and only a part of them are largely overlapped with $I$. In general, only the most pertinent input views are useful for the inter-viewpoint aggregation.
	
	To this end, we develop a view selection strategy to choose two reference views $\{I_1^r,I_2^r\}$ from the input views that are most overlapped with the rendered view $I$. Specifically, we formulate the view selection problem based on the pinhole camera model. An arbitrary 3D scene can be roughly approximated as a bounding sphere in Fig.~\ref{fig:vm}, and cameras are placed around it to take pictures. When camera-emitted rays hit the sphere, there are a set of intersections. We refer to the 3D point sets as $\Phi_i=\{p_{0}^i,p_{1}^i,\cdots,p_{M_i}^i\}$ and $\Phi_j=\{p_{0}^j,p_{1}^j,\cdots,p_{M_j}^j\}$ for the $i$-th and $j$-th cameras. For $m_i$-th intersection $p_{m_i}^i \in \Phi_i$ of view $i$, we search its nearest point in view $j$ with the L2 distance
	\begin{equation} % \{p_{0}^j,p_{1}^j,\cdots,p_{M_j}^j\}
		p_{m_i}^{i \rightarrow j} = \mathop{\arg\min}_{p \in \Phi_j} ( || p - p_{m_i}^i||_2^2 ).
	\end{equation}
	Then the matching cost from the $i$-th view to the $j$-th view is calculated by
	\vspace{-0.1in}
	\begin{equation}
		C_{i \rightarrow j} = \sum_{m_i=0}^{M_i}|| p_{m_i}^{i} - p_{m_i}^{i \rightarrow j}||_2^2 .
	\end{equation}
	We finally obtain the mutual matching cost between views $i$ and $j$ as
	\begin{equation}
		C_{i \leftrightarrow j} = C_{i \rightarrow j} + C_{j \rightarrow i}.
		\label{eq:mutualcost}
	\end{equation}
	In this regard, two reference views $\{I_1^r,I_2^r\}$ are selected at the least mutual matching costs for enhancing the NeRF-rendered view $I$. Note that we also adopt this strategy to decide the two reference views for the LLFF-T~\cite{mildenhall2019local} data during the training phase.

	\section{Experiments}
	\label{sec:exp}
	
	\subsection{Implementation Details}
	% We perform on-the-fly image degradation through the proposed NeRF-style degradation simulator on the collected dataset. To train our inter-viewpoint mixer, we adopt random $128 \times 128$ cropping, vertical or horizontal flipping, and $90^{\circ}$ rotation augmentations. Meanwhile, we also add random offsets~($\pm$2 pixels) on the reference frames to model simple camera movements. Combined with the real captured displacements between multiple frames, it is capable of covering more complex viewpoint changes. The inter-viewpoint mixer is trained for 300K iterations with a batch size of 16. We use an Adam optimizer and set the learning rate decays from $5 \times 10^{-4}$ to $0$ by a cosine annealing strategy. For efficiency, we use two~($T=2$) reference high-quality frames for inter-viewpoint aggregation. A single NeRFLiX model is trained to refine all the NeRF models for quantitative evaluation.
	We train the IVM for 300K iterations. The batch size is 16 and the patch size is 128. We adopt random cropping, vertical or horizontal flipping, and rotation augmentations. Apart from the inherent viewpoint changes over $\{I,I_1^r,I_2^r\}$, random offsets~($\pm5$ pixels) are globally applied to the two reference views~($I_1^r,I_2^r$) to model more complex motion. We adopt an Adam~\cite{kingma2014adam} optimizer and a Cosine annealing strategy to decay the learning rate from $5 \times 10^{-4}$ to 0. We train a single IVM on the LLFF-T and Vimeo datasets and test it on all benchmarks~(including user-captured scenes).
	
	\subsection{ Datasets and Metrics} 
	We conduct the experiments on three widely used datasets, including LLFF~\cite{mildenhall2019local}, Tanks and Temples~\cite{knapitsch2017tanks}, and Noisy LLFF Synthetic.
	
	% \noindent\textbf{LLFF~\cite{mildenhall2019local}} LLFF is a real-world captured dataset, where the posed information is calculated by COLMAP~\cite{schonberger2016structure}. It has eight different scenes with 20 to 62 captured images. Following previous works~\cite{tensorf, fridovich2022plenoxels, wang2021nerf, attal2022learning, Niemeyer2021Regnerf}, we consider two evaluation protocols: \textit{LLFF-P1} uses the image resolution of $1008\times 756$, and \textit{LLFF-P2} employs the image size of $504\times 376$. 
	\vspace{0.05in}
	\noindent\textbf{LLFF~\cite{mildenhall2019local}.} LLFF is a real-world dataset, where 8 different scenes have 20 to 62 images. Following the commonly used protocols~\cite{tensorf, fridovich2022plenoxels, wang2021nerf, attal2022learning, Niemeyer2021Regnerf}, we adopt $1008\times 756$ resolution for \textit{LLFF-P1} and $504\times 376$ resolution for \textit{LLFF-P2}.
	
	\vspace{0.05in} % 
	\noindent\textbf{Tanks and Temples~\cite{knapitsch2017tanks}.}
	\label{tanks}
	% It contains 5 scenes of real objects captured by an inward-facing camera circling the
	% scene. Each scene contains 152-384 images of size $1920\times 1080$. The viewpoints of different frames are much larger than LLFF~\cite{mildenhall2019local}. 
	It contains 5 scenes captured by inward-facing cameras. There are 152-384 images in the $1920\times 1080$ resolution. It should be noted that the viewpoints of different frames are significantly larger than LLFF.
	
	\vspace{0.05in}
	\noindent\textbf{Noisy LLFF Synthetic~\cite{mildenhall2020nerf}.}
	% It contains 8 virtual scenes. Each scene contains 400 images of size 800 × 800. We manually apply random camera jetting on those accurate camera poses to model noise calibration in real-world scenarios.
	There are 8 virtual scenes, each of which has 400 images with a size of $800 \times 800$. To simulate noisy in-the-wild calibration, we ad hoc apply camera jittering~(random rotation and translation are employed) to the precise camera poses.
	
	%	\paragraph{In-the-wild Scenes}
	%	We also collect some real-world scenes captured by a mobile phone for visual analysis. The last two datasets contain some simple geometry that few objects are placed. In contrast, we aims to capture such scenes with multiple objects. The captured frames contain rich and challenging image textures. So, any inaccurate predictions of NeRF models that will lead to significant visual artifacts, facilitating the observation and evaluation.
	
	% Following the previous NeRF approaches, we adopt PSNR~(${\color{red}\uparrow}$) and SSIM~(${\color{red}\uparrow}$)~\cite{wang2004image} as the quantitative evaluation metrics. The higher values indicate better results. We also use the LPIPS~(${\color{red}\downarrow}$)~\cite{zhang2018unreasonable} as a perception metric for evaluation~(the lower the better).
	\vspace{0.05in}
	\noindent\textbf{Metrics.}
	Following previous NeRF methods, we adopt PSNR~(${\color{red}\uparrow}$)/SSIM~\cite{wang2004image}~(${\color{red}\uparrow}$)/LPIPS~\cite{zhang2018unreasonable}(${\color{red}\downarrow}$) for evaluation.

	\begin{figure*}[t]
		\centering	%\vspace{-0.15in}
		\includegraphics[width=1.0\linewidth]{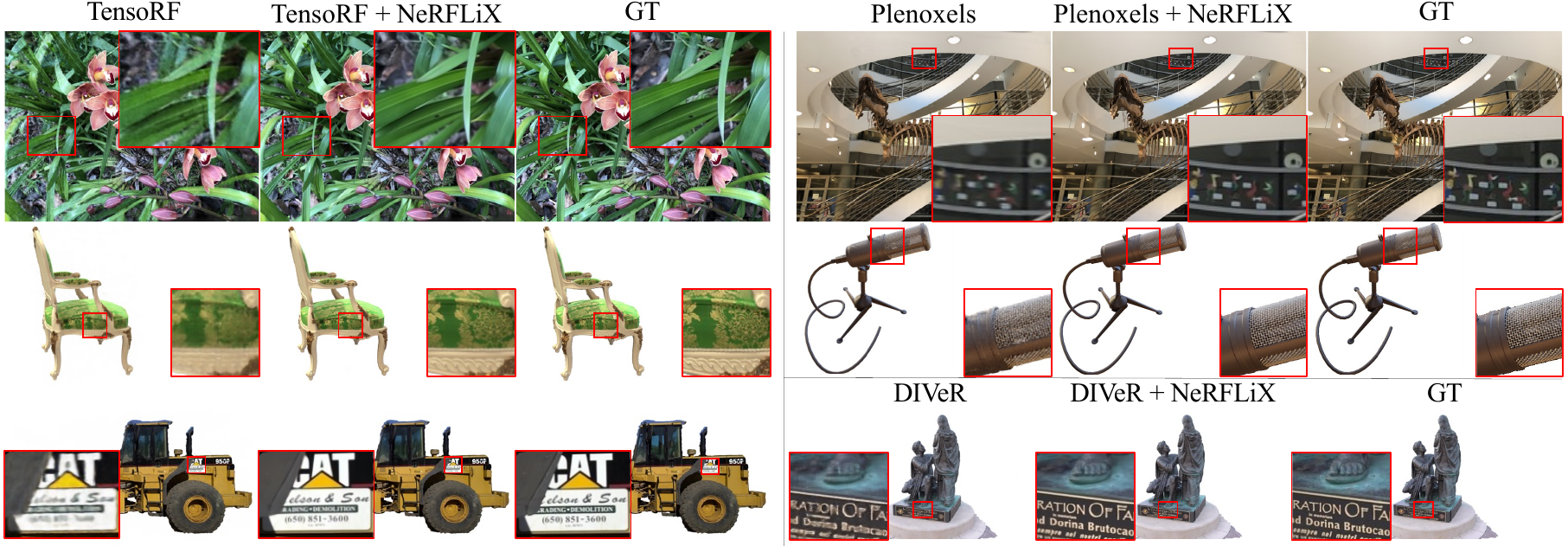} 
		
		\caption{Qualitative evaluation of the improvement over three SOTA NeRFs on LLFF, Noisy LLFF Synthetic, and Tanks and Temples.}
		\vspace{-0.1in}
		\label{fig:sotafig}
		
	\end{figure*} % 
	\subsection{Improvement over SOTA NeRF Models}
	% Without knowing camera poses, NeRF-mm presents a unified framework for optimizing neural radiance field and camera parameters simultaneously. Compare to the same model with COLMAP calibration, it obtains nosier results. RegNeRF incorporates a regularization scheme for sparse novel-view synthesis.  Both TensoRF and Plenoxels present advanced 3D voxel representations for efficient novel-view rendering and achieve state-of-the-art results on LLFF~\cite{mildenhall2019local}. However, as discussed in Section~\ref{sec:intro}, they still follow the conventional rendering pipeline of NeRF and have the potential risks of rendering artifacts, undermining the performance of novel-view synthesis.
	We demonstrate the effectiveness of our approach by showing that it consistently improves the performance of cutting-edge NeRF approaches across various datasets.
	
	%   \noindent\textbf{LLFF}
	% We evaluate the effectiveness of our proposed method by checking the improvement over six representative NeRF models, including NeRF~\cite{mildenhall2020nerf}, TensoRF~\cite{tensorf}, Plenoxels~\cite{fridovich2022plenoxels}, NeRF-mm~\cite{wang2021nerf}, NLF~\cite{attal2022learning}, and RegNeRF~\cite{Niemeyer2021Regnerf}.   Based on their noisy outputs, we evaluate the effects of the proposed NeRFLiX by checking the improvement over those SOTA models.
	% Table~\ref{tab:sota1} reports the quantitative comparison. We observe that NeRFLiX consistently pushes the performance of all the involved NeRF models to entirely new heights, under both two protocols. For Plenoxels, NeRFLiX brings \textbf{0.6dB/0.072/0.054} improvements in terms of PSNR/SSIM/LPIPS. 
	\vspace{0.05in}
	\noindent\textbf{LLFF.} In order to fully verify the generalization ability of our NeRFLiX, we investigate six representative models, including NeRF~\cite{mildenhall2020nerf}, TensoRF~\cite{tensorf}, Plenoxels~\cite{fridovich2022plenoxels}, NeRF-mm~\cite{wang2021nerf}, NLF~\cite{attal2022learning}, and RegNeRF~\cite{Niemeyer2021Regnerf}. Using rendered images of NeRF models as inputs to our model, we aim to further improve the synthesis quality. The quantitative results are provided in Table~\ref{tab:sota1}. We find that under both of the two protocols, our method raises NeRF model performance entirely to new levels. For example, NeRFLiX improves Plenoxels~\cite{fridovich2022plenoxels} by \textit{0.61dB/0.025/0.054} in terms of PSNR/SSIM/LPIPS.

	\begin{table}[t]
		\small
		\setlength{\tabcolsep}{3pt}
		\centering
		\begin{subtable}[b]{0.48\textwidth}
			\centering
			\begin{tabular}{l|c|c|c } %p{2cm}p{2cm}
				\hline
				Method                      & PSNR~(dB)${\color{red}\uparrow}$ &SSIM${\color{red}\uparrow}$ &LPIPS${\color{red}\downarrow}$   \\ \hline
				
				TensoRF~\cite{tensorf} {\footnotesize (ECCV'22)}              			 &  28.43      		     & 0.920  &0.142  \\ 
				TensoRF~\cite{tensorf} + NeRFLiX                 &  \textbf{28.94}($\color{red}\uparrow$ 0.51)         &   \textbf{0.930} &\textbf{0.120} \\  \hline \hline

				DIVeR~\cite{wu2021diver}  {\footnotesize(CVPR'22)}             			 &  28.16      		     & 0.913  &0.145  \\ 
				DIVeR~\cite{wu2021diver} + NeRFLiX                 &  \textbf{28.61}($\color{red}\uparrow$ 0.45)         &   \textbf{0.924} &\textbf{0.127} \\  \hline
			\end{tabular}
			
			\caption{Improvement over TensoRF and DIVeR on Tanks and Temples.}
			\vspace{-0.15in}
			\label{subtab:sotatanks}
		\end{subtable}
		\hfill
		\vspace{0.02in}
		\begin{subtable}[b]{0.48\textwidth}
			
			\centering
			\begin{tabular}{l|c|c|c } %p{2cm}p{2cm}
				\hline
				Method                      & PSNR~(dB)${\color{red}\uparrow}$ &SSIM${\color{red}\uparrow}$ &LPIPS${\color{red}\downarrow}$   \\ \hline
				
				TensoRF~\cite{tensorf} {\footnotesize (ECCV'22)}               			 &  22.83     		     & 0.881  &0.147  \\ 
				TensoRF~\cite{tensorf} + NeRFLiX                 &  \textbf{24.12} ($\color{red}\uparrow$ 1.29)        &   \textbf{0.913} &\textbf{0.092} \\  \hline \hline

				\hline

				Plenoxels~\cite{fridovich2022plenoxels}  {\footnotesize(CVPR'22)}           			 &  23.69      		     & 0.882  &0.127  \\ 
				Plenoxels~\cite{fridovich2022plenoxels} + NeRFLiX                 &  \textbf{25.51} ($\color{red}\uparrow$ 1.82)        &   \textbf{0.920} &\textbf{0.084} \\  \hline
			\end{tabular}
			\caption{Improvement over TensoRF and Plenoxels on noisy LLFF Synthetic.}
			\vspace{-0.15in}
			\label{subtab:sotalego}
		\end{subtable}
		\hfill
		\vspace{0.02in}
		\begin{subtable}[b]{\columnwidth}
			
			\begin{tabular}{l|c } %p{2cm}p{2cm}
				\hline
				Method                      & PSNR (dB)${\color{red}\uparrow}$/SSIM${\color{red}\uparrow}$/LPIPS${\color{red}\downarrow}$  \\ \hline
				
				TensoRF~\cite{tensorf}(4 hours)             			 &26.73/ 0.839/ 0.204  \\ 
				TensoRF~\cite{tensorf}(\textbf{2 hours})             			 &26.18/ 0.819/ 0.230  \\ 
				\cite{tensorf}(\textbf{2 hours})  + NeRFLiX                 &\textbf{27.14}/ \textbf{0.858}/ \textbf{0.165}  \\  \hline \hline

				\hline

				Plenoxels~\cite{fridovich2022plenoxels}(24 minutes)            			 &26.29/ 0.839/ 0.210  \\ 
				Plenoxels~\cite{fridovich2022plenoxels}(\textbf{10 minutes})                  &25.73/ 0.804/ 0.252\\
				
				\cite{fridovich2022plenoxels}(\textbf{10 minutes}) + NeRFLiX                 &\textbf{26.60}/ \textbf{0.847}/ \textbf{0.181} \\  \hline

				% EfficientNeRF~\cite{hu2022efficientnerf}(4 hours)            			 &  26.29      		     / 0.839  /0.210  \\ 
				% \cite{hu2022efficientnerf}(\textbf{2 hours}) + NeRFLiX                 &  \textbf{25.51} /   \textbf{0.920} /\textbf{0.084} \\  \hline
			\end{tabular}
			\caption{Improvement over TensoRF and Plenoxels trained with half of the 
				recommended iterations on LLFF~\cite{mildenhall2019local} under LLFF-P1.}
			\vspace{-0.1in}
			\label{subtab:sotahalf}
		\end{subtable}
		\caption{Quantitative evaluation of the improvement of NeRFLiX for various NeRFs.}\label{tab:sotasynthetic}
		\vspace{-0.3in}
	\end{table}
	
	% \noindent\textbf{Tanks and Temples}
	% As described in~\ref{tanks}, Tanks and Temple~\cite{knapitsch2017tanks} is a challenging in-the-wild 3D scene dataset captured with distinctive viewpoints. Even with advanced rendering representation, TensoRF~\cite{tensorf}, DIVeR~\cite{wu2021diver} show inferior synthetic quality on this dataset. Here, we demonstrate how NeRFLiX enhances the results of TensoRF and DIVeR. Table~\ref{subtab:sotatanks} shows the detailed results. We observe that NeRFLiX gets \textbf{0.51}/\textbf{0.01}/\textbf{0.022} improvements over the TensoRF in terms of PSNR/SSIM/LPIPS. Our NeRFLiX recovers sharp image details and efficiently removes NeRF-style degradations, as illustrated in Figure~\ref{fig:sotafig}(b).
	% This experiment illustrates that our NeRFLiX is capable of boosting those difficult scenarios with significant visual artifacts.
	\vspace{0.05in}
	\noindent\textbf{Tanks and Temples.} Due to large variations of camera viewpoints, even advanced NeRF models, e.g., TensoRF~\cite{tensorf} and DIVeR~\cite{wu2021diver}, show obviously inferior rendering quality on this dataset. As illustrated in Table~\ref{subtab:sotatanks}, we show that our NeRFLiX can still boost the performance of these models by a large margin, especially TensoRF~\cite{tensorf} achieves \textit{0.51dB}/\textit{0.01}/\textit{0.022} improvements on PSNR/SSIM/LPIPS. 
	% In addition, as shown in Fig.~\ref{fig:sotafig}, our method successfully removes severe rendering artifacts and produces sharper image details.

	% \noindent\textbf{Noisy LLFF Synthetic}
	% Apart from the two in-the-wild testing benchmarks, we demonstrate the enhancement ability of our method on noisy LLFF Synthetic.  The results are shown in Table~\ref{subtab:nsds2}. It can be seen that our NeRFLiX obtains substantial improvements over the two SOTA NeRFs. 
	\vspace{0.05in}
	\noindent\textbf{Noisy LLFF Synthetic.} Apart from in-the-wild benchmarks above, we also demonstrate the enhancement capability of our model on noisy LLFF Synthetic. From the results shown in Table~\ref{subtab:sotalego}, we see that our NeRFLiX yields substantial improvements for two SOTA NeRF models.

	%  \noindent\textbf{Qualitative Results}
	% Figure~\ref{fig:sotafig} depicts some visual examples for qualitative evaluation. It is evident that our NeRFLiX restores clear image details and removes most NeRF-style artifacts of those degraded images produced by SOTA NeRF models, yielding promising results.
	\vspace{0.05in}
	\noindent\textbf{Qualitative results.} In Fig.~\ref{fig:sotafig}, we provide some visual examples for qualitative assessment. It is obvious that our NeRFLiX restores clearer image details while removing the majority of NeRF-style artifacts in the rendered images, clearly manifesting the effectiveness of our method. More results are provided in our supplementary materials.

	\subsection{Training Acceleration for NeRF Models}
	%    Since NeRFLiX can restore high-quality frames from variant NeRF, we investigate its potential to improve those early terminated NeRF models. To do this, we train three SOTA NeRF models with half the recommended training iterations as in their paper. The quantitative results are presented in Table~\ref{subtab:sotahalf}. Despite only half the training time, thanks to NeRFLiX, those models still outperform their counterparts trained with complete epochs. Notably, the early terminated Plenoxels models~(only 10 minutes) refined with NeRFLiX outperform the original models, clearly demonstrating NeRFLiX's efficacy.
	In this section, we show how our approach makes it possible for NeRF models to produce better results even with a 50$\%$ reduction in training time. To be more precise, we use NeRFLiX to improve the rendered images of two SOTA NeRF models after training them with half the training period specified in the publications. The enhanced results \textit{outperform} the counterparts with full-time training, as shown in Table~\ref{subtab:sotahalf}.  Notably, NeRFLiX has reduced the training period for Plenoxels~\cite{fridovich2022plenoxels} from 24 minutes to 10 minutes while also improving the quality of the rendered images.
	
	\subsection{Ablation Study}
	% In this section, we conduct comprehensive experiments on LLFF~\cite{mildenhall2019local} under LLFF-P1 to thoroughly analyze each of our designs in this part. We use TensoRF~\cite{tensorf} as our baseline by default\footnote{Here, the TensoRF results~(26.70dB/0.838/0.204) tested by us are slightly lower than their original paper~(26.73dB/0.839/0.204).}.
	In this section, we conduct comprehensive experiments on LLFF~\cite{mildenhall2019local} under the LLFF-P1 protocol to analyze each of our designs. We use TensoRF~\cite{tensorf} as our baseline\footnote{The TensoRF results~(26.70dB/0.838/0.204) we examined differ slightly from the published results~(26.73dB/0.839/0.204).}.
	\vspace{-0.1in}
	\subsubsection{NeRF-Style Degradation Simulator}
	
	\begin{figure}[ht]
		\centering
		%\vspace{-0.15in}
		\includegraphics[width=0.95\columnwidth]{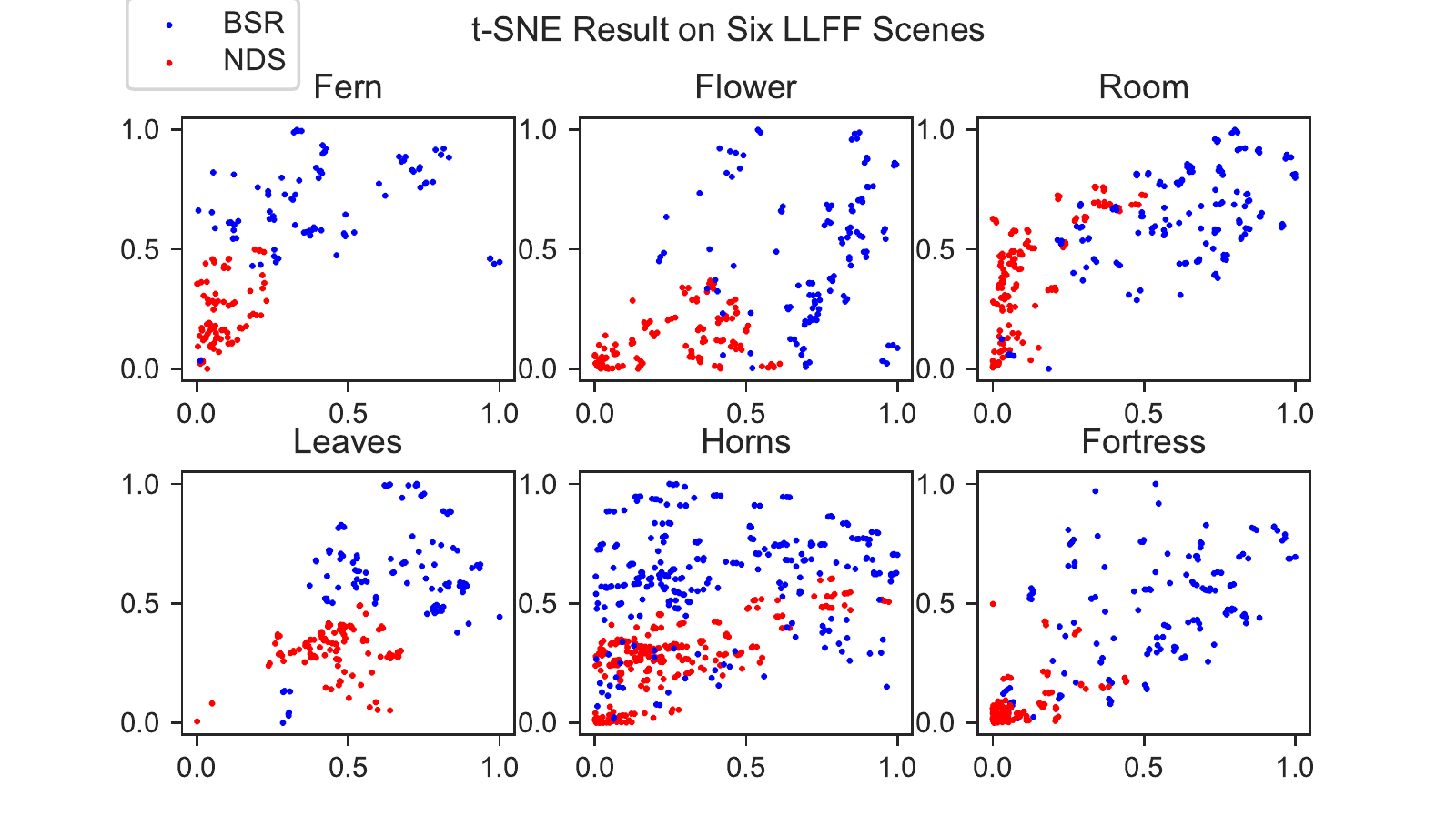} 
		
		\caption{ Quantitative comparison between our NDS and BSR~\cite{zhang2021designing} over six LLFF scenes. We draw the normalized differences between the simulated images of the two degradation methods and the real NeRF-rendered images. The smaller values, the better results are (best viewed in color).
		} 
		\vspace{-0.3in}
		\label{fig:tsne}
	\end{figure} % 
	
	\vspace{0.05in}
	\noindent\textbf{Simulation quality.} 
	% We first examine the simulation quality of our proposed NeRF-style degradation simulator~(NSDS).  To this end, we conduct a quantitative comparison with the BSR degradation~\cite{zhang2021designing}. We collect the whole results and their corresponding ground truths of two SOTA NeRF models on LLFF~\cite{mildenhall2019local}. Then, taking the noise-less ground truths as inputs, we employ BSR degradations and our NeRF-style degradations in Secion~\ref{sec:degraded} to obtain degraded images, respectively. We use t-SNE to visualize deep image features~(by inception-v3) of those image sets. The results are illustrated in Figure~\ref{fig:tsne}. With systemically NeRF-style degradation analysis, our simulated dataset is statistically closer to the real rendered frames than BSR~\cite{zhang2021designing}, indicating the simulation superiority of our degradation designs.
	We first examine the simulation quality of the proposed NeRF-style degradation simulator. To this end, we analyze the distribution of our degraded images, BSR~\cite{zhang2021designing} degraded images and NeRF rendered images on LLFF~\cite{mildenhall2019local}. We use t-SNE~\cite{van2008visualizing} to visualize deep image features~(by Inception-v3~\cite{szegedy2016rethinking}) and results are shown in Fig.~\ref{fig:tsne}. Our simulated data is statistically much closer to the real rendered images than BSR. This conclusion is also supported by Table~\ref{subtab:nsds2}, which demonstrates that our NDS significantly surpasses BSR and yields 0.6-1.0dB improvements when used for learning NeRF degradations.

	\begin{table}[t]
		\small

		\begin{subtable}[b]{0.46\textwidth}
			\centering
			\setlength{\tabcolsep}{2pt}
			\begin{tabular}{c|c|c|c|c|c } %p{2cm}p{2cm}
				\hline
				Model &TensoRF(Base)  & SwIR{$\mathrm{\bf _N}$}   &DATSR{$\mathrm{\bf _N}$}              & EDVR{$\mathrm{\bf _N}$}   &VST{$\mathrm{\bf _N}$}  \\ \hline
				
				PSNR  &26.70      &26.82   &26.84               	&26.88  &26.79    		      \\
				SSIM  &0.838      &0.845 &0.843                &0.847   &0.842  \\
				\hline 
				%						Models                  &\cite{tensorf} + SWIR + BSRGAN  &\cite{tensorf} + SWIR + Ours   \\ \hline
				%						Metrics             & 26.19/0.834   	  &26.82/0.845      		      \\
				%						\hline \hline
				
			\end{tabular}
			
			\caption{Quantitative results of the improvements using existing image/video processing models trained on our simulated dataset.}
			\label{subtab:nsds1}
		\end{subtable}
		\hfill
		\begin{subtable}[b]{0.46\textwidth}
			
			\begin{tabular}{l|c|c|c|c|c|c } %p{2cm}p{2cm}
				\hline
				Models  & BSR & NDS & SwIR & IVM &PSNR &SSIM\\ \hline
				SwIR$\bf{_B}$        &\cmark  &      &\cmark & & 26.20& 0.834 \\
				SwIR{$\mathrm{\bf _N}$}       &  &\cmark      &\cmark & & \textbf{26.82}& \textbf{0.845} \\  \hline
				IVM$\bf{_B}$         &\cmark  &      & &\cmark & 26.40& 0.842 \\
				IVM{$\mathrm{\bf _N}$}         &  &\cmark      & &\cmark & \textbf{27.39}& \textbf{0.867}\\ \hline

			\end{tabular}
			\caption{Comparison of our NDS and the BSR degradation models~\cite{zhang2021designing}.}
			\label{subtab:nsds2}
		\end{subtable}
		\caption{Quantitative comparison between our NDS and BSR~\cite{zhang2021designing}. We use subscripts $\bf{_N}$, $\bf{_B}$, to represent the models trained with our NDS dataset and BSR, respectively.}\label{tab:payoffMatrices}
		
	\end{table}

	\vspace{0.05in}
	\noindent\textbf{Degradation type.}
	% In addition, we also examine the influence of each degradation type in our NeRF-style degradation simulator. Apart from our full model that adopts all the involved degradations, we gradually increase those degradations and train another three models for comparison. Table~\ref{table:diffnoises} shows the quantitative results on LLFF~\cite{mildenhall2019local}. It concludes that each of the employed degradations contributes to the final performance.
	We also evaluate the detailed contribution of each data degradation. We use simulated data to train our models by gradually including four types of degradation, as illustrated in Table~\ref{table:diffnoises}. From quantitative comparisons on LLFF~\cite{mildenhall2019local}, we observe that all employed degradations are beneficial to our system.
	
	% \noindent\textbf{Simulated Dataset.}  As depicted in Table~\ref{subtab:nsds1}, the models, including DATSR~\cite{cao2022datsr}, SwIR~\cite{liang2021swinir}, EDVR~\cite{wang2019edvr},  VST~\cite{liu2022video}, trained on our proposed simulated dataset~(with \textbf{a lower latter ``$_N$"}) deliver better quantitative results than the baseline~(26.70/0.838). Additionally, Table~\ref{subtab:nsds2} shows that the models trained using the proposed NeRF-style degradation simulator~(NDS) significantly surpass their counterparts with BSR degradation modeling~\cite{zhang2021designing}~(with \textbf{a lower latter ``$_B$"}). 
	\begin{table}[t]
		\small
		\setlength{\tabcolsep}{3pt}
		% \ding{52}
		\begin{center}
			
			\begin{tabular}{c|c|c|c|c|c|c } %p{2cm}p{2cm}
				\hline
				Models                      &SGN &Re-Pos. &A-Blur &RA  &PSNR(dB) &SSIM \\ \hline
				
				Model-1             		&\cmark	 &       		     &   & &27.08&0.856  \\ 
				Model-2                 	&\cmark	 &\cmark       		     &   & &27.13&0.858  \\
				Model-3             		&\cmark	 &\cmark       		     &\cmark   & &27.21&0.859  \\ 
				Model-4                 	&\cmark	 &\cmark       		     &\cmark   &\cmark &\textbf{27.39}&\textbf{0.867}  \\
				\hline
				
				\hline
				
			\end{tabular}
		\end{center}
		\vspace{-0.2in}
		\caption{Influences of different degradations used in our NeRF-style degradation simulator. ``SGN" and ``RA" are shorted for splatted Gaussian noise and region-adaptive schemes and ``A-Blur" refers to anisotropic Gaussian blur. }
		\label{table:diffnoises}
	\end{table}

	\begin{figure}[t]
		\centering
		%\vspace{-0.15in}
		\includegraphics[width=1.0\columnwidth]{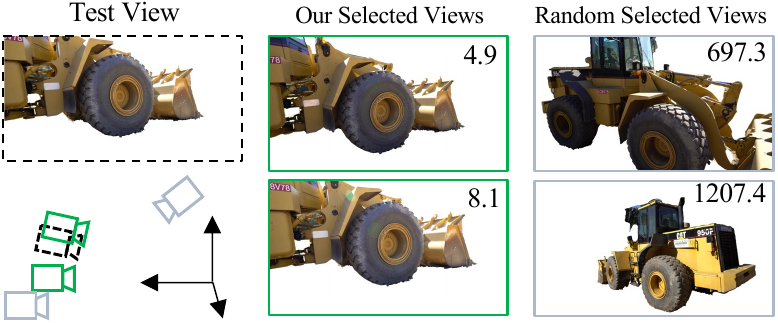} 
		
		\caption{Comparison of the two view selection methods. The number in each sub-figure is the matching cost calculated in Eq.~\ref{eq:mutualcost}. 
		} 
		
		\label{fig:vmviz}
	\end{figure} % 

	\vspace{-0.1in}
	\subsubsection{Inter-viewpoint Mixer}
	\noindent\textbf{View selection strategy.}
	% We also check the impact of our view-matching strategy. To this end, apart from this strategy, we also adopt random training frame selection for inter-viewpoint fusion. In Figure~\ref{fig:vmviz}, we show some selected frames by the two different methods. As can be seen, our view-matching method collects the most relevant reference training images for IVM aggregation. In addition, we also quantitatively evaluate the influence of our proposed view-matching approach. As shown in Table~\ref{table:viewmatching}, the results obtained by the proposed view-matching strategy achieve better performance, confirming the effects of our view-matching design.
	We develop a view selection strategy to make full use of high-quality reference views. As shown in Fig.~\ref{fig:vmviz}, our system can identify the most relevant views for quality enhancement when compared to random selection. Also, the quantitative results in Table ~\ref{table:viewmatching} suggest that our view selection achieves significantly improved results, illustrating the usefulness of our method.
	% In Fig.~\ref{fig:vmviz}, we also quantitatively evaluate our view, selection model. The proposed view selection technique achieves significantly improved performance, illustrating the usefulness of our method.
	
	\begin{table}[t]
		\small
		\setlength{\tabcolsep}{6pt}

		% \ding{52}
		\begin{center}
			
			\begin{tabular}{c|c|c } %p{2cm}p{2cm}
				\hline
				Method                     & LLFF &Tanks and Temple  \\ \hline
				
				Random             		  &  27.06dB/ 0.856       		     &  28.51dB/ 0.925  \\ 
				View Selection        &  \textbf{27.39dB}/ \textbf{0.867}         &   \textbf{28.94dB}/ \textbf{0.930}  \\   \hline
			\end{tabular}
		\end{center}
		\vspace{-0.2in}
		\caption{Ablation studies of our view selection strategy. }
		\vspace{-0.2in}
		\label{table:viewmatching}
		
	\end{table}
	
	\vspace{0.05in}
	\noindent\textbf{Hybrid recurrent multi-view aggregation.}
	% In this work, we develop a hybrid recurrent inter-viewpoint aggregation network for NeRF-rendered image restoration. Here, we assess the effects of our aggregation module. Apart from the model trained with the proposed aggregation strategy, we also train its variant counterparts that adopt different aggregation schemes. As shown in Table~\ref{table:align}, those models use either pixel-wise or patch-wise aggregation strategy, obtaining inferior results than our proposed hybrid recurrent aggregation method. 
	% Additionally, it costs an extra 66-46 ms per iteration while achieving 0.11/0.06dB gains by gradually raising the iteration numbers (from R1 to R3).
	To handle large viewpoint differences in reference and rendered views, we develop a hybrid recurrent inter-viewpoint aggregation network. We train models using either pixel-wise or patch-wise aggregation and test different iterations to assess the proposed IVM. Models using a single aggregation approach, as illustrated in Table~\ref{table:align}, perform worse than our full configuration. Additionally, by gradually increasing the iteration numbers from 1 to 3, we achieve improvements of 0.12dB and 0.06dB, albeit at an additional cost of 66 ms and 46 ms for aggregation. Last, compared with the existing state-of-the-art models in Table~\ref{subtab:nsds1}, thanks to the recurrent hybrid aggregation strategy, our IVM outperforms all of these models in terms of quantitative results, demonstrating the strength of our aggregation design.
	
	% \noindent\textbf{Model Architecture} 
	% % Here, we illustrate how cutting-edge deep neural works, including EDVR~\cite{wang2019edvr}, SwIR~\cite{liang2021swinir}, VST~\cite{liu2022video}, may be utilized to eliminate NeRF visual artifacts using the simulated dataset. As depicted in Table~\ref{subtab:nsds1}, the models trained on our proposed simulated dataset deliver better quantitative results than the baseline. Additionally, Table~\ref{subtab:nsds2} shows that the models trained using the proposed NeRF-style degradation simulator~(NSDS) significantly surpass their counterparts with BSR degradation modeling~\cite{zhang2021designing}. 
	% Taking advantage of the simulated dataset, we can use a variety of deep neural networks to eliminate NeRF rendering artifacts. We compare backbones based on a single image, multiple reference images, and video frames, including SwinIR~\cite{liang2021swinir}, MASA(citation), EDVR~\cite{wang2019edvr} and VST~\cite{liu2022video}. As depicted in Table~\ref{subtab:nsds1}, our IVM outperforms all other baselines in terms of quantitative results, demonstrating the strength of our model.
	
	\begin{table}[t]
		\small
		\setlength{\tabcolsep}{6pt}

		% \ding{52}
		\begin{center}
			
			\begin{tabular}{c|c|c|c|c } %p{2cm}p{2cm}
				\hline
				Method                      & PSNR(dB) &SSIM & LPIPS  &Speed (ms) \\ \hline
				
				Pixel-wise             		 &     27.13   &0.862   	&0.179 	     					& 230\\ 
				Patch-wise                   &      27.00   &0.854    	&0.183		& 237\\   \hline
				Hybrid + R1                  &    27.21       & 0.865   & 0.173& 181 \\
				Hybrid + R2                  &     27.33      & 0.866    & 0.157& 247\\
				Hybrid + R3                  &  \textbf{27.39}&\textbf{0.867}         &\textbf{0.149}   &293  \\   \hline
			\end{tabular}
		\end{center}
		\vspace{-0.2in}
		\caption{Ablation studies of hybrid inter-viewpoint aggregation module. The running time is tested with an input size of $256\times 256$. }
		\label{table:align}
		\vspace{-0.15in}
	\end{table}
	
	% \subsubsection{More Adjacent Frames}
	
	% \noindent\textbf{Visualization of the Aggregation Features}
	
	\vspace{0.05in}
	\noindent\textbf{Limitation.} Though NeRFLiX achieves promising progress for its universal improvements over existing NeRF models. There are still some future directions that deserve further exploration.
	(1) Our NDS is one of many possible solutions for NeRF degradation simulation.
	(2) Exploring real-time inter-viewpoint mixers is interesting and useful.
	
	\section{Conclusion}
	We presented NeRFLiX, a general NeRF-agnostic restoration paradigm for high-quality neural view synthesis. We systematically analyzed the NeRF rendering pipeline and introduced the concept of NeRF-style degradations. Towards to eliminate NeRF-style artifacts, we presented a novel NeRF-style degradation simulator and constructed a large-scale simulated dataset. Benefiting from our simulated dataset, we demonstrated how SOTA deep neural networks could be trained for NeRF artifact removal. To further restore missing details of NeRF rendered frames, we proposed an inter-viewpoint mixer that is capable of aggregating multi-view frames captured from free viewpoints. Additionally,  we developed a view selection scheme for choosing the most pertinent reference frames, largely alleviating the computing burden while achieving superior results. Extensive experiments have verified the effectiveness of our NeRFLiX. Code will be made publicly available.
	
	%%%%%%%%% REFERENCES
	{\small
		\bibliographystyle{ieee_fullname}
		\bibliography{NeRFLiX}
	}

	\begin{appendices}
		\section{NeRF Degradation Simulator}
		\vspace{0.05in}
		\noindent\textbf{Raw data collection.}
		We collect raw sequences from Vimeo90K~\cite{xue2019video} and LLFF-T~\cite{mildenhall2019local}. In total, Vimeo90K contains 64612 7-frame training clips with a $448\times256$ resolution. Three frames~(two reference views and one target view) are selected from a raw sequence of Vimeo90K in a random order. As described in Sec.~{\color{red}5.1}, apart from the inherent displacements within the selected views, we add random global offsets to the two reference views, largely enriching the variety of inter-viewpoint changes.
		On the other hand, we also use the training split of the LLFF dataset, which consists of 8 different forward-facing scenes with 20-62 high-quality input views. Following previous work, we drop the eighth view and use it for evaluation. To construct a training pair from LLFF-T, we randomly select a frame as the target view and then use the proposed view selection algorithm~(Sec.~{\color{red}4.3}) to choose two reference views that are most overlapped with the target view. 
		% On the other hand, we randomly select three frames from a 7-frame sequence of Vimeo90k. One frame is the target view and the others are the two reference views. 
		% For the raw sequence from Vimeo90K, as described in Sec.~{\color{red}5.1}, apart from the inherent displacements between a rendered view and its two reference views, we add random global offsets to the two reference views, largely enriching the variety of inter-viewpoint changes.
		\begin{figure}[t]
			\centering
			%\vspace{-0.15in}
			\includegraphics[width=0.8\columnwidth]{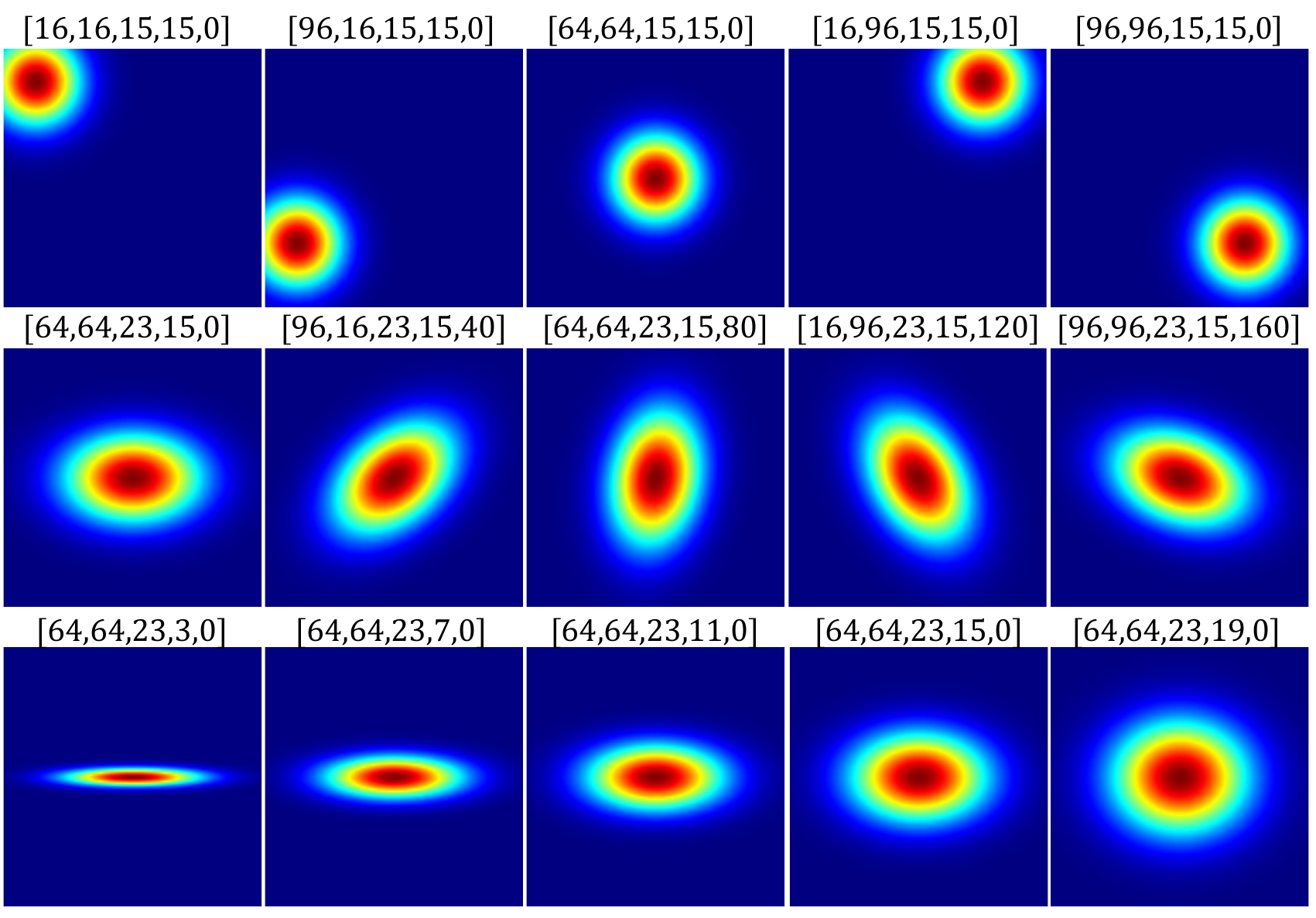} 
			
			\caption{We give some visualized region-adaptive masks. The parameters refer to the values of $[c_i, c_j; \sigma_i, \sigma_j, A]$ in Eq.~({\color{red}3}).
			} 
			\label{fig:ra}
			\vspace{-0.15in}
		\end{figure} % 
		
		\vspace{0.05in}
		\noindent\textbf{Hyper-parameter setup.}
		In Eq.~({\color{red}1}), the 2D Gaussian noise map $n$ is generated with a zero mean and a standard deviation ranging from 0.01 to 0.05. The isotropic blur kernel $g$ has a size of $5 \times 5$. We employ a Gaussian blur kernel to produce blurry contents by randomly selecting kernel sizes~(3-7), angles~(0-180), and standard deviations~(0.2-1.2). Last, in order to obtain a region-adaptive blending map $M$ in Eq.~({\color{red}3}), we use random means~($c_i,c_j \in(-16,144)$), standard deviations~($\sigma_i \in (13,25),\sigma_j \in(0,24)$), and orientation angles~($A \in(0,180)$). Additionally, we visualize some generated masks using different hyper-parameter combinations~($[c_i, c_j; \sigma_i, \sigma_j, A]$) in Fig.~\ref{fig:ra}.
		
		\vspace{0.05in}
		\noindent\textbf{Training data size.}
		We investigate the influence of training data size. Under the same training and testing setups, we train several models using different training data sizes. As illustrated in Table~\ref{table:DATA}, we can observe that the final performance is positively correlated with the number of training pairs. Also, we notice the IVM trained with only LLFF-T data or additional few simulated pairs~(10$\%$ of the Vimeo90K) fails to enhance the TensoRF-rendered results, \ie, there is no obvious improvement compared to TensoRF~\cite{tensorf}. This experiment demonstrates the importance of sizable training pairs for training a NeRF restorer.
		
		\begin{table}[t]
			\small
			\setlength{\tabcolsep}{5pt}
			% \ding{52}
			\begin{center}
				
				% \begin{tabular}{c|c|c|c|c|c } %p{2cm}p{2cm}
				% 	\hline
				% 	Method           &TensoRF (Base)          &L &L-V10 &L-V50 &L-V100 \\ \hline
				
				% 	PSNR (dB)    &26.70     &26.28        &26.71&27.08                                           & \textbf{27.39} \\ 
				% 	SSIM    &0.838      &0.837              &0.840&0.856                                              &\textbf{0.867}\\ \hline
				% \end{tabular}
				\begin{tabular}{l|c|c|c|c|c } %p{2cm}p{2cm}
					\hline
					Settings  & 10$\%$ & 50$\%$ & 100$\%$ &PSNR~(dB)&SSIM\\ \hline
					LLFF-T       & & & &26.28& 0.837 \\
					LLFF-T+       &\cmark & & &26.71& 0.840 \\
					LLFF-T+       & &\cmark & &27.08& 0.856 \\
					LLFF-T+       & & &\cmark &\textbf{27.39}&\textbf{ 0.867} \\ \hline \hline 
					TensoRF (Base)  &- &- &- &26.70 &0.838 \\ \hline

				\end{tabular}
			\end{center}
			\vspace{-0.2in}
			\caption{Quantitative results of different training data sizes. First, we train an IVM model only using the LLFF-T. Then, we gradually increase the simulated pairs~(10$\%$,  50$\%$, 100$\%$) from Vimeo90K~\cite{xue2019video} to train another three IVM models. }
			\vspace{-0.15in}
			\label{table:DATA}
			
		\end{table}
		\begin{figure*}[t]
			\begin{center}
				\includegraphics[width=\linewidth]{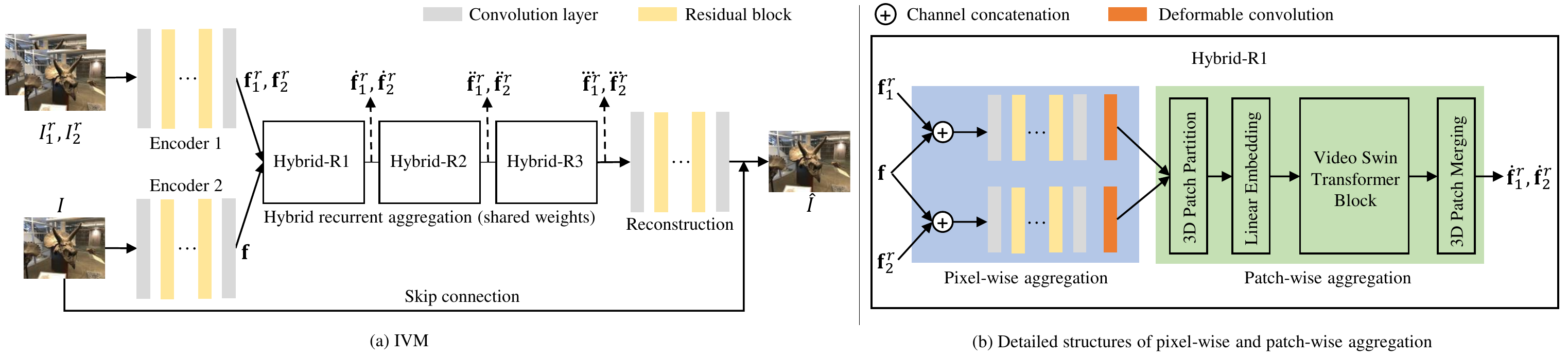} 
			\end{center}
			\vspace{-0.25in}
			\caption{The detailed framework architecture of our proposed IVM.
			}
			\label{fig:framework}
			\vspace{-0.15in}
		\end{figure*}
		\vspace{-0.05in}
		\section{Inter-viewpoint Mixer}
		\vspace{0.05in}
		In Sec.~{\color{red}4.2}, we briefly describe the framework architecture of our inter-viewpoint mixer (IVM). Here we provide more details. As illustrated in Fig.~\ref{fig:framework}(a), there are two convolutional modules~(``Encoder 1/2") to extract features of the degraded view $I$ and its two reference views $\{I_1^r,I_2^r\}$, respectively. Then, we develop a hybrid recurrent aggregation module that iteratively performs pixel-wise and patch-wise fusion. At last, a reconstruction module is implemented by a sequence of residual blocks~(40 blocks) to output the enhanced view ${\hat I}$. The default channel size is 128.
		
		\vspace{0.05in}
		\noindent\textbf{Feature extraction.}
		Given a rendered view $I$ and its two reference views ${I}_{1,2}^r$, we aim to utilize the two encoders to extract deep image features $\mathbf{f}$ and $\mathbf{f}_{1,2}^r$, respectively. As detailed in Fig.~\ref{fig:framework}(a), the two encoders share an identical structure. A convolutional layer is first adopted to convert an RGB frame to a high-dimensional feature. Then we further extract the deep image feature using 5 stacked residual blocks followed by another convolution layer. 
		% The two encoders share an identical structure. A convolutional layer is first used to convert a RGB frame to a high-dimensional feature. Then we extract the deep image feature using 5 stacked residual blocks followed by another convolution layer. to obtain the deep image features $\mathbf{f}$ and $\mathbf{f}_{1,2}^r$ from the rendered view $I$ and two reference views ${I}_{1,2}^r$, respectively.
		
		\vspace{0.05in}
		\noindent\textbf{Hybrid recurrent aggregation.} As depicted in Fig.~\ref{fig:framework}(a), we employ three hybrid recurrent aggregation blocks~(termed ``Hybrid-R1(2,3)") to progressively fuse the inter-viewpoint information from the image features~($\mathbf{f}$ and $\mathbf{f}_{\{1,2\}}^r$). Next, we take the first iteration as an example to illustrate our aggregation scheme.
		
		\vspace{0.05in}
		\noindent\textbf{Pixel-wise aggregation.}
		As shown in Fig.~\ref{fig:framework}(b), we first merge the target view feature $\mathbf{f}$ and one of the reference features~$\mathbf{f}_{\{1,2\}}^r$ by channel concatenation. Then we use a convolutional layer to reduce the channel dimension and five residual blocks followed by another convolutional layer to obtain a fused deep feature. Later on, the fused feature and the reference feature are further aggregated through a deformable convolution. And the other reference image follows the same processing pipeline. In this case, we finally obtain two features after the pixel-wise aggregation.
		
		\vspace{0.05in}
		\noindent\textbf{Patch-wise aggregation.}
		We adopt a window-based attention mechanism~\cite{liu2022video} to accomplish patch-wise aggregation. In detail, the pixel-wisely fused features are first divided into several 3D slices through a 3D patch partition layer. Then, we obtain 3D tokens via a linear embedding operation and aggregate patch-wise information using a video Swin transformer block. Finally, 3D patches are regrouped into a 3D feature map.
		
		In the next iteration, we split the 3D feature map into two ``reference" features $\mathbf{\dot{f}}_{\{1,2\}}^r$ and repeat the pixel-wise and patch-wise aggregation. Note that, the weights of pixel-wise and patch-wise modules are shared across all iterations to reduce the model complexity.
		
			\noindent\textbf{Comparisons with SOTA image restorers.}
			% Thanks for the good advice. In table 3b, we compare the \textit{officially} trained BSRGAN model to its counterpart using our simulated dataset. The latter model surpasses the officially released model by 0.62dB. Here, we try to use multiple off-the-shell image deblurring and denoising models~(Restormer CVPR-2022, MPRNet CVPR-2021, RealBasicVSR CVPR-2022) to remove the NeRF-rendered artifacts and report the results in the Table below. It is obvious that neither of the two best combinations can enhance the NeRF-rendered frames. 
%			Thanks for this good suggestion. We have tested five off-the-shelf SOTA image/video restoration models (``M1-5" refer to BSRGAN~\cite{zhang2021designing} in ICCV 2021, MPRNet~\cite{zamir2021multi} in CVPR 2021, Real-ESRGAN~\cite{wang2021realesrgan} in ICCVW 2021, Restormer~\cite{zamir2022restormer} and RealBasicVSR~\cite{chan2022investigating} in CVPR 2022, respectively) to remove NeRF-rendered artifacts. These methods, though good for real-world image restoration, are not designed nor trained specifically to tackle NeRF-style artifacts. In fact, as shown in the Table~\ref{tab:enmodel} below, the quality of the synthesized images is basically not improved or becomes even worse. Conversely, if using our simulated NDS dataset to re-train the BSRGAN model, we find it obviously outperforms the original BSRGAN by 0.62dB~(Table 3b of the main paper). This experiment indicates the existing image/video restoration model cannot enhance NeRF-rendered frames, confirming the necessity of NeRFLiX.
Thanks for this good suggestion. We have tested five off-the-shelf SOTA image/video restoration models (``M1-5" refer to BSRGAN in ICCV 2021, MPRNet in CVPR 2021, Real-ESRGAN in ICCVW 2021, Restormer and RealBasicVSR in CVPR 2022, respectively) to remove NeRF-rendered artifacts. These methods, though good for real-world image restoration, are not designed nor trained specifically to tackle NeRF-style artifacts. In fact, as shown in the Table~\ref{tab:enmodel} below, the quality of the synthesized images is basically not improved or becomes even worse. Conversely, if using our simulated NDS dataset to re-train the BSRGAN model, we find it obviously outperforms the original BSRGAN by 0.62dB~(Table 3b of the main paper). This experiment indicates the existing image/video restoration model cannot enhance NeRF-rendered frames, confirming the necessity of NeRFLiX.
			\vspace{-0.12in}
			\begin{table}[h]
				\small
				\centering
				\setlength{\tabcolsep}{5pt}
				\begin{tabular}{l|c||c|c|c|c|c } %p{2cm}p{2cm}
					\hline
					Methods &Base    &M1 &M2 &M3 &M4 &M5     \\ \hline
					PSNR (dB) &26.70 &26.20 &26.35 &25.43 &26.24 &25.03  \\
					SSIM &0.838 &0.834 &0.825 &0.814 &0.822 &0.795  \\ \hline
					% PSNR	  &26.70 &26.24    &26.35 &25.03 &26.20 &\textbf{26.82} \\ 
					%           SSIM	     &0.838 &0.822   &0.825 &0.795 &0.834 &\textbf{0.845} \\
					
				\end{tabular}
				\caption{Quantitative results of five image/video restoration models, where ``Base" mode refers to TensoRF.}
				\label{tab:enmodel}
				\vspace{-0.2in}
			\end{table}
		
		\section{Additional Results}
		
		\begin{table}[t]
			\small
			\setlength{\tabcolsep}{6pt}

			% \ding{52}
			\begin{center}
				
				\begin{tabular}{c|c|c|c|c } %p{2cm}p{2cm}
					\hline
					Method                     & IVM-0V & IVM-1V &IVM-2V &IVM-3V \\ \hline
					
					PSNR (dB)        &26.87        &27.26&    27.39                                           &27.44\\ 
					SSIM        &0.846              &0.862&0.867                                              &0.869\\ \hline
				\end{tabular}
			\end{center}
			\vspace{-0.2in}
			\caption{Quantitative results of different numbers of reference views. }
			\vspace{-0.15in}
			\label{table:vn}
			
		\end{table}
		\vspace{0.05in}
		\noindent\textbf{Number of reference views.}
		By default, we perform inter-viewpoint aggregation using two reference views (termed IVM-2V). We train another three models (IVM-0V, IVM-1V, and IVM-3V) adopting different numbers of reference views. The results are shown in Table~\ref{table:vn}. The model without using reference views~(IVM-0V) achieves the lowest PSNR and SSIM values compared with other models. Meanwhile, it is observed that the more reference views used, the higher IVM performance, indicating the importance of utilizing high-quality reference views.
		% The models without reference views or one yield 0.52 and 0.19 PSNR decreases compared to IVM-2V. Additionally, including more reference views is able to further improve the performance.
		% , while requiring the extra cost of ** for processing more neighboring frames.
		
\begin{figure*}[t]
			\begin{center}
				\includegraphics[width=1.0\linewidth]{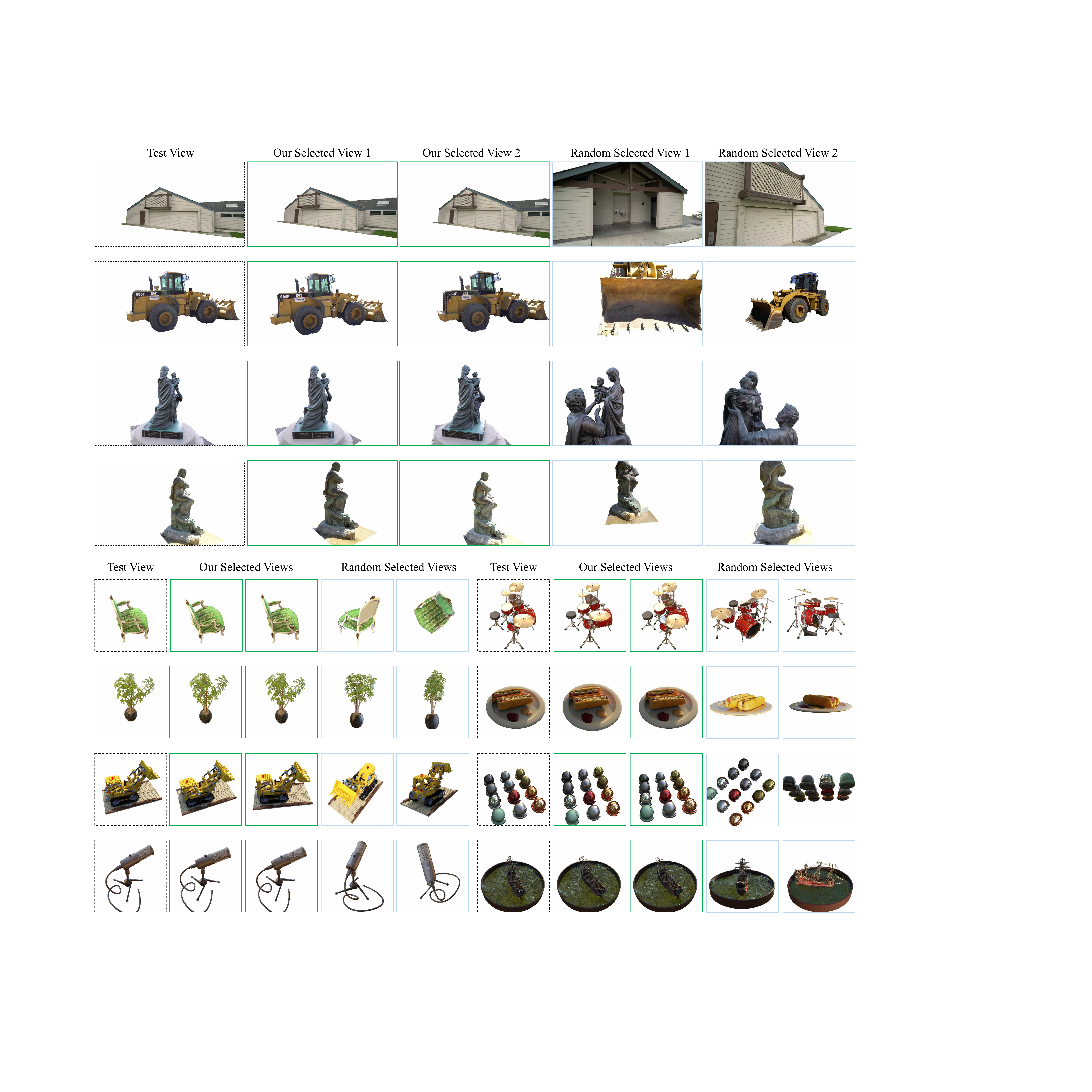} 
			\end{center}
			\vspace{-0.15in}
			\caption{Visual comparison between two view selection methods.
			}
			\label{fig:vs}
		\end{figure*}

		\vspace{0.05in}
		\noindent\textbf{View selection.}
		Fig.~\ref{fig:vs} exhibits the selected views by our algorithm in different NeRF scenes. We see that the proposed view selection strategy is able to choose the most relevant ones from freely captured views.
		
\begin{figure*}[t]
			\begin{center}
				\includegraphics[width=\linewidth]{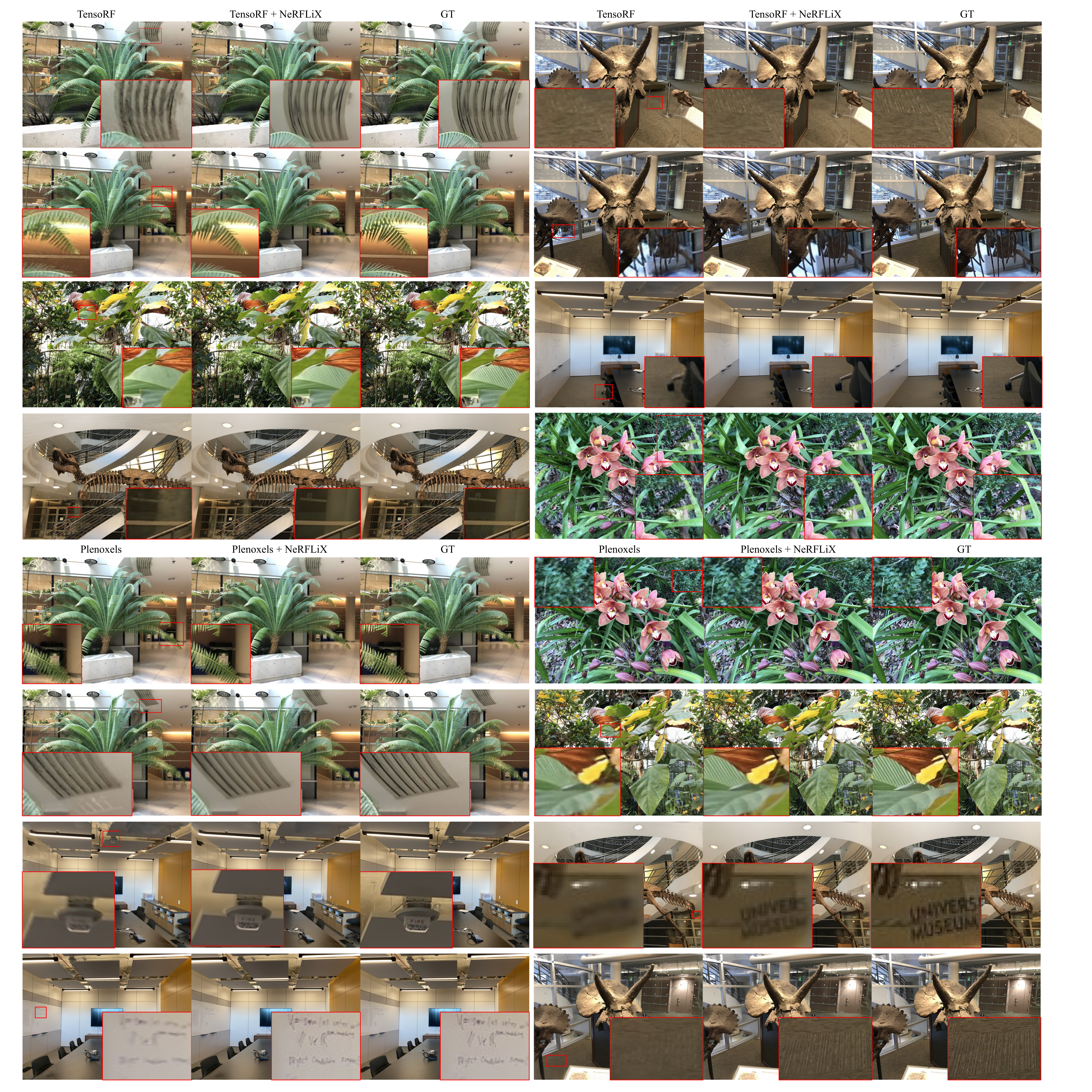} 
			\end{center}
			%\vspace{-0.15in}
			\caption{Qualitative evaluation of the improvement over two SOTA NeRF models~(TensoRF~\cite{tensorf} and Plenoxels~\cite{fridovich2022plenoxels}) on LLFF~\cite{mildenhall2019local} under LLFF-P1.
			}
			\label{fig:sotallff}
		\end{figure*}

\begin{figure*}[t]
			\begin{center}
				\includegraphics[width=\linewidth]{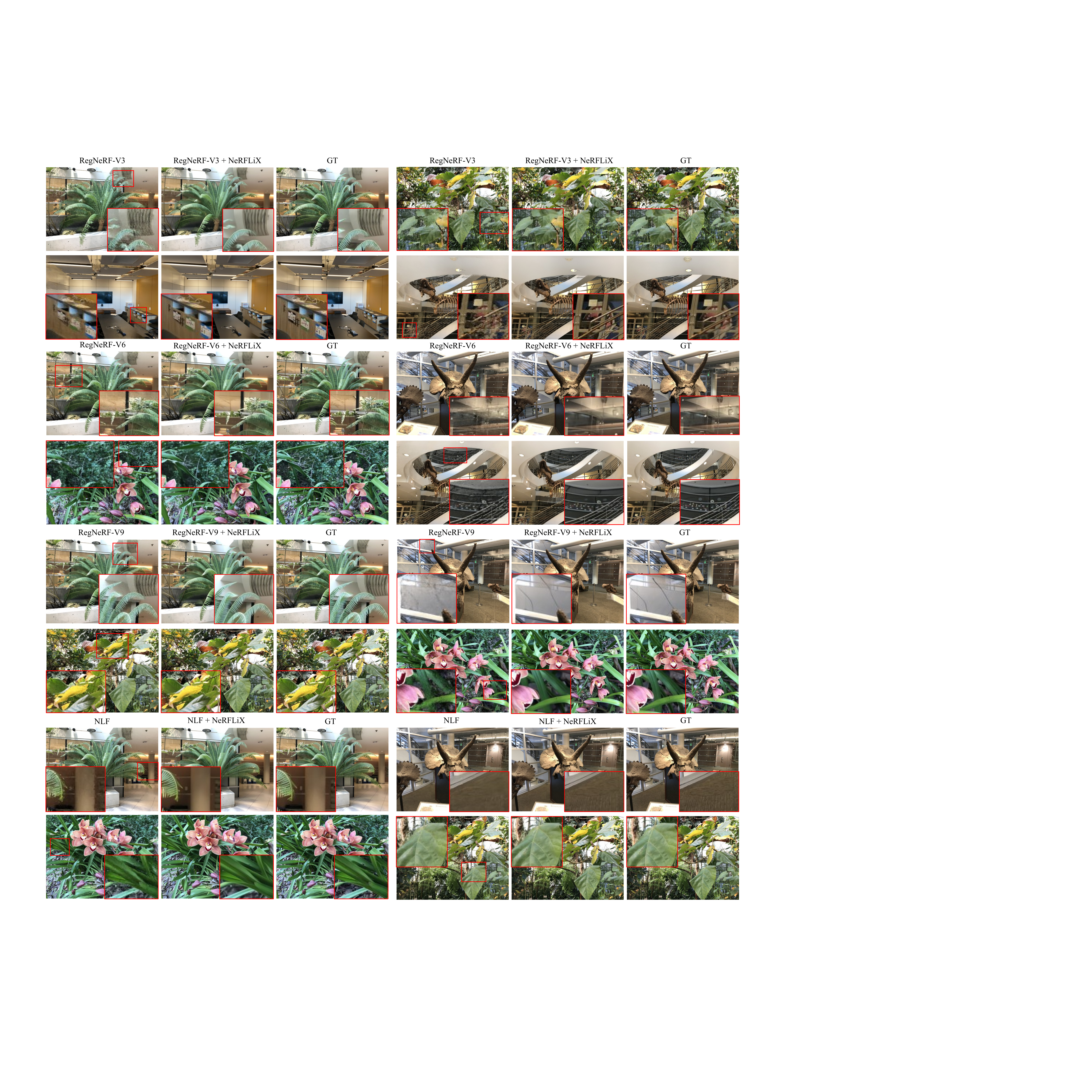} 
			\end{center}
			%\vspace{-0.15in}
			\caption{Qualitative evaluation of the improvement over two SOTA NeRF models~(RegNeRF~\cite{Niemeyer2021Regnerf} and NLF~\cite{attal2022learning}) on LLFF~\cite{mildenhall2019local} under LLFF-P2. RegNeRF-V3(6,9) takes 3(6,9) input views for training.
			}
			\label{fig:sotallff1}
		\end{figure*}

\begin{figure*}[h]
			\centering
			\begin{subfigure}{\linewidth}
				\centering
				\includegraphics[width=0.8\linewidth]{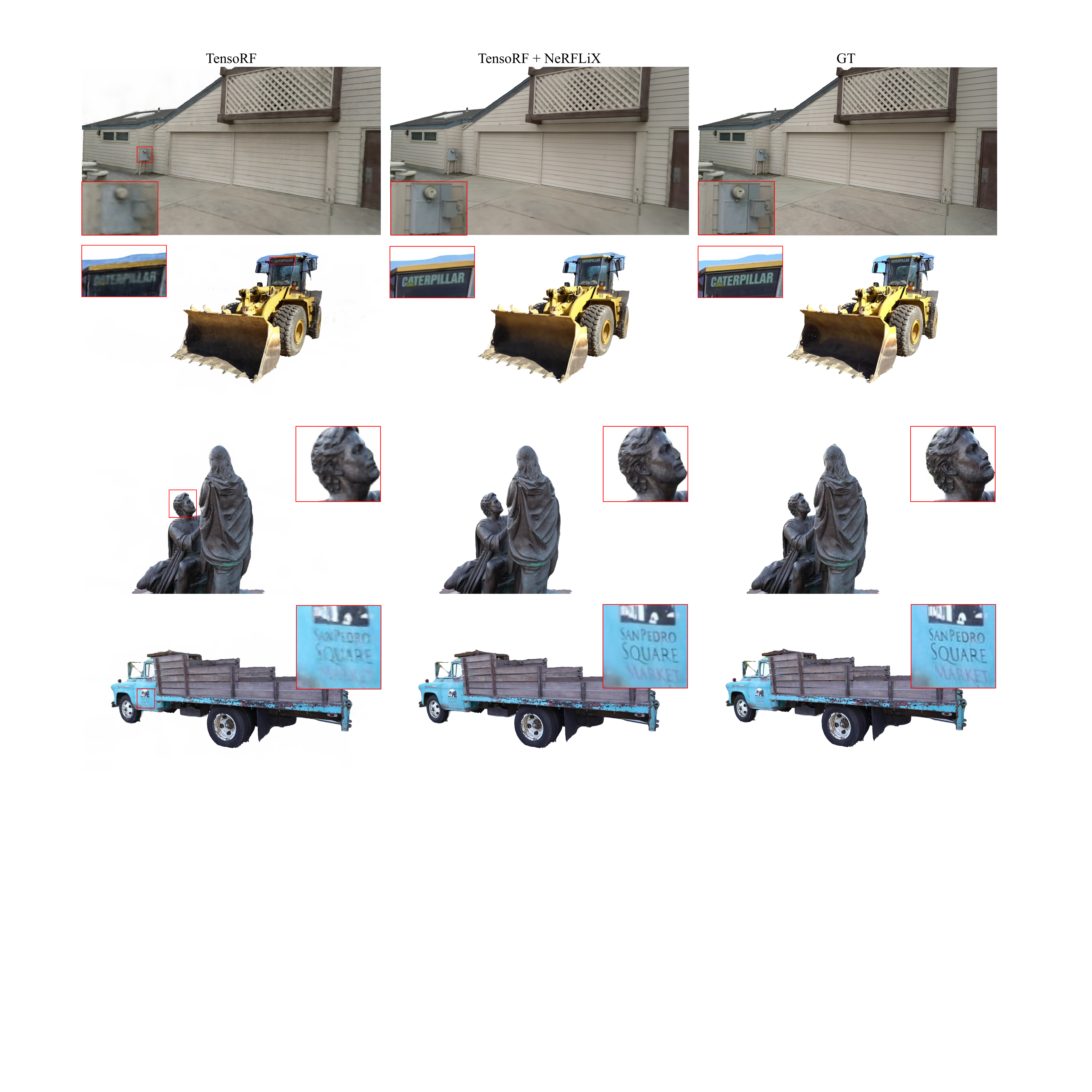}  % 
				
				\caption{Qualitative evaluation of the improvement over TensoRF~\cite{tensorf} on Tanks and Temples~\cite{knapitsch2017tanks}. } 
				\label{fig:tanka}
				
			\end{subfigure}
			\hfill
			\begin{subfigure}{\linewidth}
				\centering
				\includegraphics[width=0.8\linewidth]{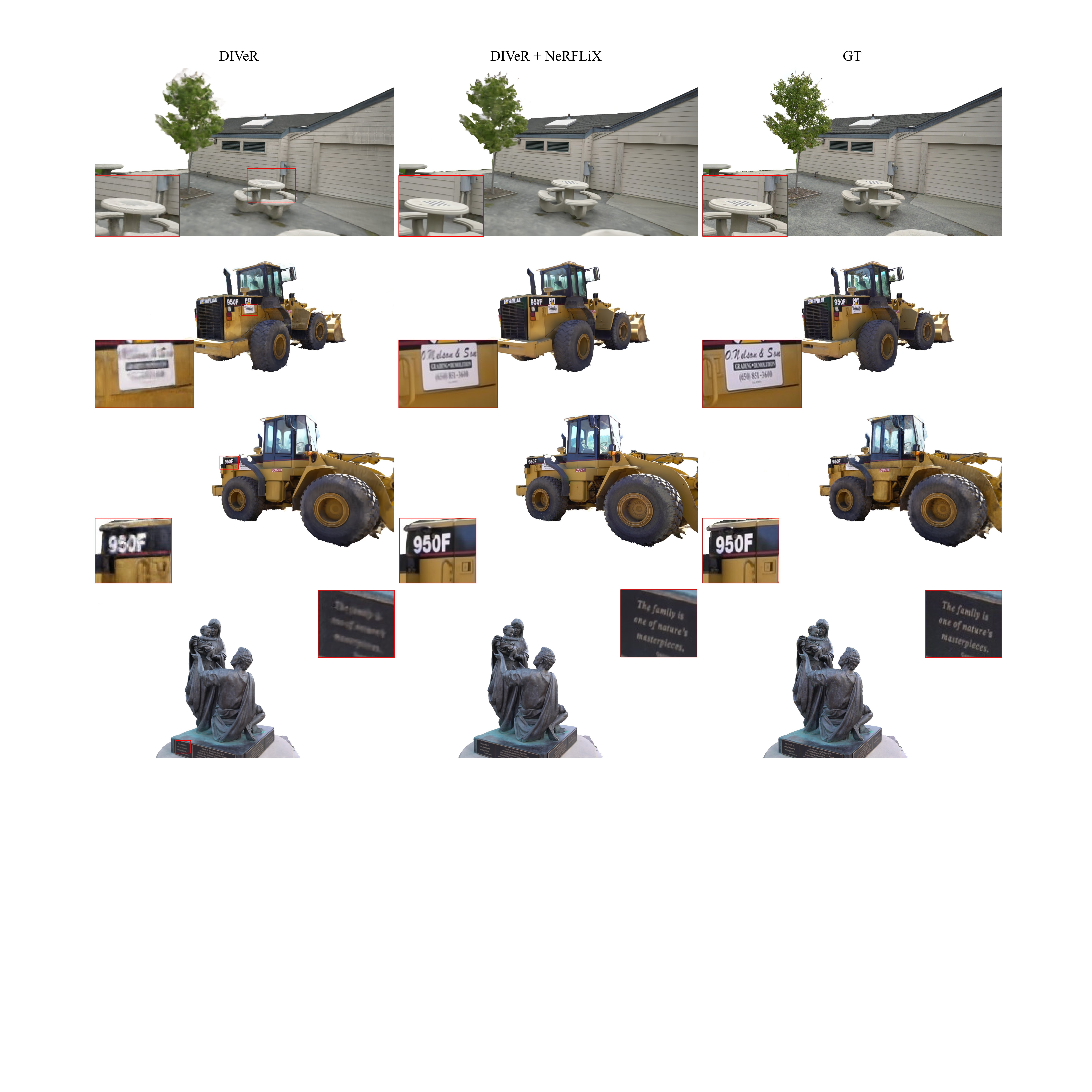}  % 
				
				\caption{Qualitative evaluation of the improvement over DIVeR~\cite{wu2021diver} on Tanks and Temples~\cite{knapitsch2017tanks}.  } 
				\label{fig:tankb}
			\end{subfigure}
			\caption{Qualitative evaluation of the improvement over two SOTA NeRF models on Tanks and Temples~\cite{knapitsch2017tanks}.  } 
			\label{fig:tanks}
			\vspace{-0.1in}
		\end{figure*}

\begin{figure*}[t]
			\begin{center}
				\includegraphics[width=\linewidth]{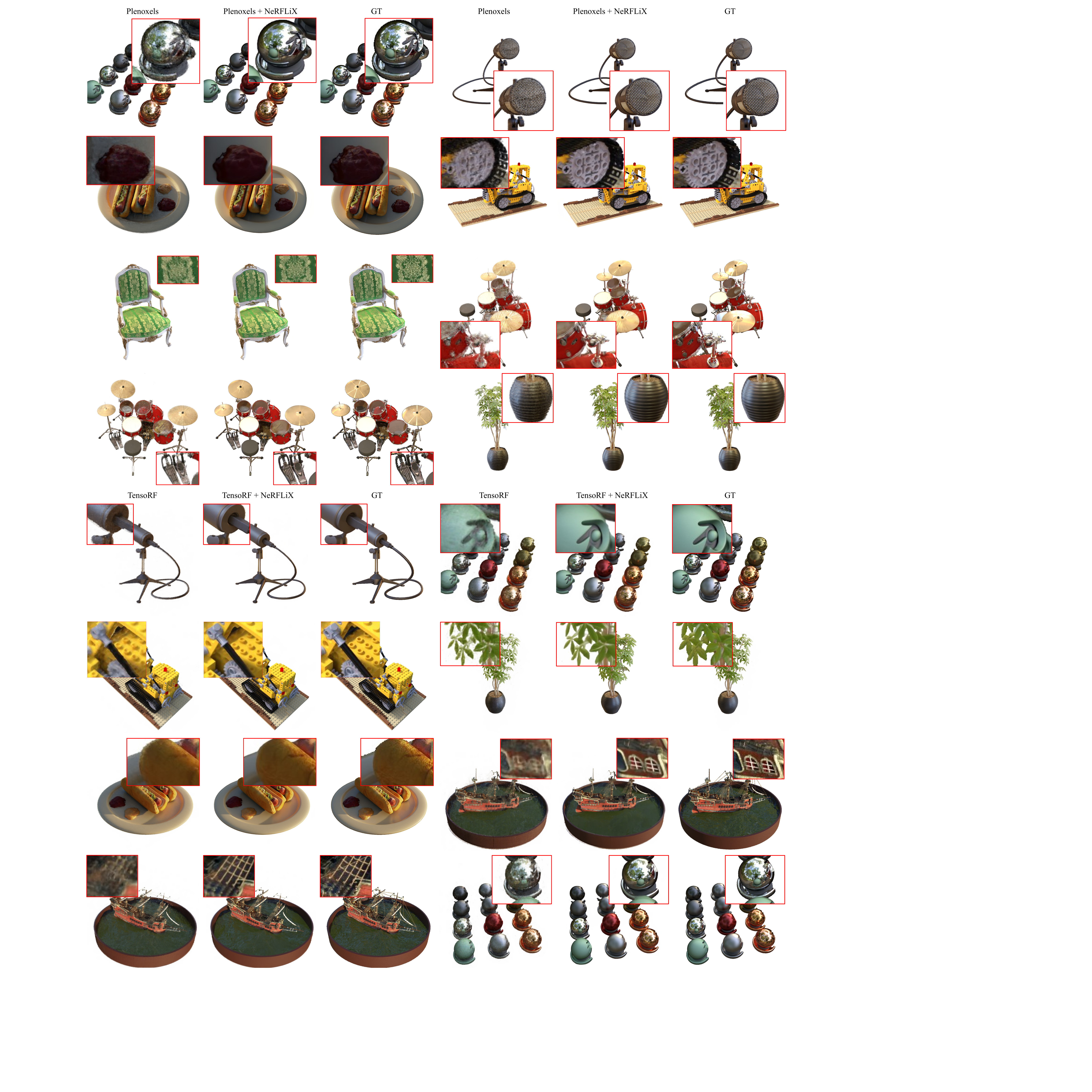} 
			\end{center}
			%\vspace{-0.15in}
			\caption{Qualitative evaluation of the improvement over two SOTA NeRF models~(Plenoxels~\cite{fridovich2022plenoxels} and TensoRF~\cite{tensorf}) on noisy LLFF Synthetic.
			}
			\label{fig:lego}
		\end{figure*}

		\vspace{0.05in}
		\noindent\textbf{Qualitative results.}
		%In our paper, extensive experiments demonstrate our proposed NeRFLiX consistently improves NeRF-rendered quality, pushing the performance of cutting-edge NeRF models to entirely new levels. Notably, even with noisy camera parameters, fewer training iterations(time), and relatively limited input views~(3 input views + RegNeRF + NeRFLiX), our model shows impressive quantitative and qualitative improvements, indicating the effectiveness of NeRFLiX. Last but not the least, we show that a pioneering, earlier NeRF model, referred to as ~\cite{mildenhall2020nerf} achieves state-of-the-art performance.
		Here, we provide more visual examples to adequately validate the effectiveness of our approach. As shown in Fig.~\ref{fig:sotallff}, Fig.~\ref{fig:sotallff1}, Fig.~\ref{fig:tanks}, Fig.~\ref{fig:lego}, NeRFLiX consistently improves NeRF-rendered images with clearer details and fewer artifacts for all NeRF models.  For example, NeRFLiX successfully recovers recognizable characters, object textures, and more realistic reflectance effects, while effectively eliminating the rendering artifacts.
		
		\vspace{0.05in}
		\noindent\textbf{Video demo.}
		We also provide a video demo\footnote{available at our project website \url{https://redrock303.github.io/nerflix/}} for a clear visual comparison. First, we show some NeRF-rendered views and the restored counterparts of NeRFLiX. Then, we provide two video cases (one is from LLFF and the other is an in-the-wild scene) to compare the rendered views of TensoRF~\cite{tensorf} and enhanced results of our NeRFLiX. It is observed that NeRFLiX is capable of producing clearer image details and removing the majority of the NeRF rendering artifacts.  
		

	\end{appendices}
\end{document}